# Design Challenges of Multi-UAV Systems in Cyber-Physical Applications: A Comprehensive Survey, and Future Directions

Reza Shakeri, Mohammed Ali Al-Garadi, Ahmed Badawy, Amr Mohamed, Tamer Khattab, Abdulla Al-Ali,  Khaled A. Harras, Mohsen Guizani

*Abstract*— Unmanned Aerial Vehicles (UAVs) have recently rapidly grown to facilitate a wide range of innovative applications that can fundamentally change the way cyber-physical systems (CPSs) are designed. CPSs are a modern generation of systems with synergic cooperation between computational and physical potentials that can interact with humans through several new mechanisms. The main advantages of using UAVs in CPS application is their exceptional features, including their mobility, dynamism, effortless deployment, adaptive altitude, agility, adjustability, and effective appraisal of real-world functions anytime and anywhere. Furthermore, from the technology perspective, UAVs are predicted to be a vital element of the development of advanced CPSs. Therefore, in this survey, we aim to pinpoint the most fundamental and important design challenges of multi-UAV systems for CPS applications. We highlight key and versatile aspects that span the coverage and tracking of targets and infrastructure objects, energy-efficient navigation, and image analysis using machine learning for fine-grained CPS applications. Key prototypes and testbeds are also investigated to show how these practical technologies can facilitate CPS applications. We present and propose state-of-the-art algorithms to address design challenges with both quantitative and qualitative methods and map these challenges with important CPS applications to draw insightful conclusions on the challenges of each application. Finally, we summarize potential new directions and ideas that could shape future research in these areas.

## I. INTRODUCTION

MILLIONS of Unmanned Aerial Vehicles (UAVs), also known as drones, are estimated to be massively adopted in various real-life applications, from civilian (surveillance, transportation, environmental monitoring, industrial monitoring, agriculture services, and disaster relief) [1] to military services (air exploration, battlefield surveillance, target localization, tracking, damage assessment, and anti-terrorism arrests) [1]. The popularity of UAVs is attributed to their vital features, such as mobility, adaptive altitude, effortless deployment, flexibility, and adjustable usage. Practical applications that use UAVs to gather sensor data are significant because they utilize the advantages of wireless sensor networks (WSNs), such as low costs, efficient sensors with no deployment overhead, accessibility to diverse data types, including temperature, humidity, and geographic information [1]. Similar to WSNs, the application scenarios that can benefit from UAV technologies are diverse, countless in numbers, and can have versatile social, economic, and environmental impacts in numerous industry domains. In general, almost all applications leveraging WSN technologies can be re-engineered to leverage UAV technologies either as a replacement or a complementary solution. Moreover, since UAVs sensors are versatile and mobile, these sensors can be deployed on demand.

UAVs are useful in many crucial industrial applications for the non-destructive testing and pre-sensing of catastrophic failures [2]. Some of these important industries are oil flare stacks, wind turbines, power transmission towers, solar collectors, photovoltaic or solar panels, steel structures, and high-rise buildings. In such contexts, UAVs can help collect thermographic and visual images, in addition to other types of data for use in numerical simulations. For example, air enters flare stacks during operation because unpredictable blockages of the off-gas flow can result in a catastrophic explosive mixture of air and hydrocarbons in the stacks [3]. Similarly, in numerous cases, setting up fixed cameras or constantly using human inspection is not feasible, a drawback that emphasizes the need for UAV systems. [1]

Surveillance and remote sensing are amongst the top industries for monitoring human behavior and changes in the environment with the objectives of public safety and security as well as infrastructure protection. Video-based surveillance market is expected to reach  over 68 billion dollars by 2023 [4]. Technologies like satellite remote sensing (SRS) and distributed camera sensor networks (CSNs) are used in many applications to provide large scale and localized monitoring,

Reza Shakeri, Mohammed Ali Al-Garadi, Amr Mohamed, Abdulla Al-Ali, and Mohsen Guizani are with Department of Computer Science and Engineering, Qatar University, 2713, Doha, Qatar. E-mails: shakeri.reza.work@gmail.com, {mohammed.g ,amrm, abdulla.alali}@qu.edu.qa  mguizani@ieee.org. Ahmed

Badawy and Tamer Khattab are with Department Electrical Engineering, Qatar University, Doha, Qatar E-mails: badawy234@gmail.com , tkhattab @ieee.org. Khaled A. Harras  is with Carnegie Mellon University Qatar Campus, 92971 Doha, E-mails: kharras@cs.cmu.edu.

respectively. However, as shown in Fig.1, both technologies provide two extremes for surveillance with respect to accuracy versus intrusive deployment and maintenance. SRS provides slow large-scale monitoring, while CSNs require widespread deployment and constant human intervention for setup and maintenance, which may not be feasible particularly in harsh environments. On the one hand, SRS cannot be used for fine-grain monitoring, such as monitoring of structural objects, with high accuracy, which limits its usage in many applications that require close monitoring and context detection [5]. On the other hand, traditional sensor networks (SNs) are spreading in countless number of Internet of Things (IoT) applications to provide automatic monitoring of objects, humans, and/or infrastructure. SNs, however, require permanent deployment of large magnitude of localized sensors, intrusively to the environment in many cases, and also require constant human intervention to manage and maintain. The proliferation of low-cost UAVs [6] is emerging to provide innovative ways of low altitude sensing with zero-deployment (i.e., no fixed deployment requirements) that can fundamentally change the way such applications are designed.

Similar to surveillance applications, UAV technologies have progressed significantly in several applications in the past decade, making commercial and civil applications feasible and compelling. As it occurs with many disruptive technologies, UAVs owe their development to the defense industry. However, recently, there was a widespread adoption of this technology for civilian use. As a result, the market for commercial/civilian UAVs is expected to grow at a compound annual growth rate of 19% between 2015 and 2020, compared with 5% growth on the military side. The interest in research and innovative solutions leveraging this technology has recently grown exponentially, witnessed by the number of publications and patents that became available in this area [7]. UAVs can provide cost-effective bio-inspired surveillance solutions to monitor humans and objects closely. In this way, facilitating the detection of complex events such as border intrusions [8], pipeline failures, public security, and vegetation analysis [9]. Other UAV applications include automatic detection of road regions [10], mobile relaying [11] , [12] and relaying for wireless physical layer security [13]. For further details on civil applications of UAVs along with their communication network requirements such as quality of service and data requirements, we refer the reader to [8], [14] and the references therein.

Cyber Physical systems (CPSs) are promising new systems that combine both virtual and physical worlds by a synergy of computational power and physical processes. The core of CPSs is the integration of ubiquitous computing systems and communication services. Within the paradigm of CPSs, an interface between sensors, devices, network users and/or applications either with each other or with their surroundings is systematized by communication and computation cores [15]. Applications of CPSs are enormous including medical systems, manufacturing automation, transportation and aviation systems [16].

The following subsections introduce UAV systems, CPSs, and potential applications of Multi-UAVs in CPSs. These applications will be used in the remainder of this survey to map and identify the design challenges of UAVs with the CPSs applications. Subsection D discusses related surveys and highlights the key contributions of this survey.

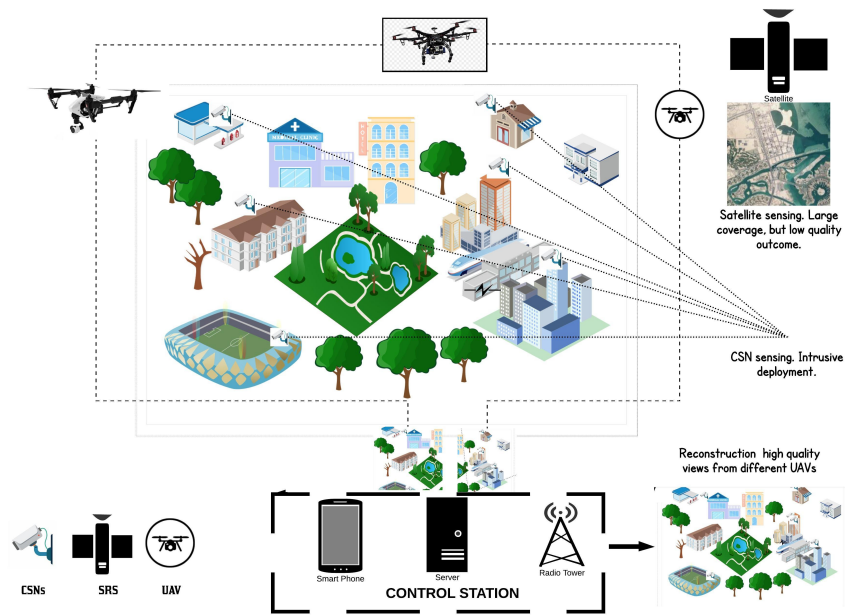

Fig.1. CSNs, SRS and UAV sensing systems

*A. UAV Systems*

As shown in Fig. 2, UAVs can be categorized into different categories; based on their size into micro or Nano UAVs, small UAVs, medium UAVs, and large UAVs. Based on their flying range, UAVs are classified into very close range (approximately 5 km radius), close range (approximately 50 km), short range (approximately 150 km), mid-range (approximately 650 km), and large range (greater than 650 km). With regard to their altitude, UAVs are classified into low (less than approximately 330 m), medium (approximately 9,000 m), and high (20,000 m or more) UAVs. UAVs are also grouped based on their endurance, namely, long flight and short flight durations. Moreover, UAVs are classified based on their applications, namely, civil (e.g., surveillance, transportation, environmental monitoring, and industrial monitoring) and military (e.g., target localization, tracking, and anti-terrorism arrests). They are also grouped based on their type, namely, fixed wing and rotary wing. These UAV classifications are based on previous researches [4, 17, 18].

In addition to the development of embedded systems and the miniaturization tendency of micro-electro-mechanical systems, manufacturing small or mini UAVs at low cost has become feasible [19]. However, a single UAV system can only provide limited operational tasks to achieve full operational function and provide widespread application of the system; the effective collaboration and synchronization of multiple UAVs can construct a system that uses effectively UAVs compared with a single UAV. The advantages and disadvantages of both multiple and single UAVs are shown in Table 1.

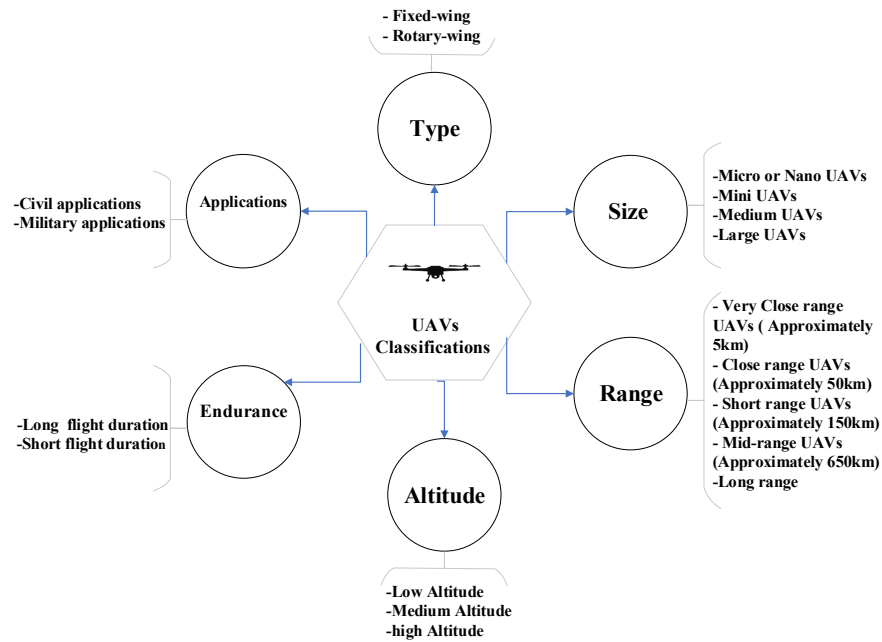

Fig.2. UAVs classifications

TABLE 1
ADVANTAGES AND DISADVANTAGES OF MULTI-UAV AND SINGLE-UAV SYSTEMS

| Comparison features | Multi-UAV systems | Single-UAV systems |
|---|---|---|
| **Targeted area coverage** | Multiple UAVs can cover larger areas compared with single UAVs. | Single UAVs can only cover smaller areas compared with multi-UAVs [20]. |
| **Cost** | Maintenance costs of a mini multi-UAV are less than the cost of a large UAV [19, 21]. | Large UAVs may not require high maintenance costs. |
| **Task time** | The required task can be completed faster with multi-UAVs [19]. | Completion of the required task is slower with single UAVs. |
| **Radar cross-section** | Requires a smaller radar cross-section [19, 22] | Requires a larger radar cross-section |
| **Power** | Since Multi-UAV system is constructed using small UAVs, UAVs in such systems are power-constrained compared with single large UAV systems. | The power in single large UAV systems is efficient compared with that in mini multi-UAVs. |
| **Network topology** | Complex network topology requirements | Requires direct and simple network connection |
| **Application** | Wide range of applications | Limited range of applications |
| **Security** | Multiple UAV systems have large attack surfaces and vectors, particularly when they are part of an Internet of Things (IoT) system. | Generally, single UAV systems are less vulnerable compared to multi-UAV systems. However, they provide single point of failure in case of successful attacks. |

When it comes to monitoring an area of interest, it is obvious that one drone is typically not adequate to cover and monitor every point within that area unless the area is too small which is not commonly found in practical scenarios. We focus on the investigation of multi-UAV in practical CPS application scenarios. Under this assumption, there are several issues that need to be addressed effectively and properly, including but not limited to, area/target coverage; UAV path planning; and networking and energy considerations, amongst others. In this survey, we are interested in highlighting such design aspects of multi-UAV systems for CPS applications. The coordination and formation in *3D* space amongst several flying objects, each having specific responsibility, that result in a unified goal will open the door for new challenges. Hence, many considerations should be considered to guide the crowd of UAVs to efficiently execute their assigned duties and avoid unexpected behavior. For example, there is a trade-off between minimizing the time required to cover all targets in a specific area, while minimizing the number of UAVs engaged in target coverage. In other words, we can use one drone to cover very large areas in the simplest form, using the concept of mobile camera, which will be very time consuming. Alternatively, we can use large number of UAVs deployed in strategic places to cover all key targets in the area in a very short time. Another trade-off lies in how to control the mobility of UAVs for mobile target tracking such that all targets are covered on every instance of time, while minimizing the energy consumed in mechanical movements, leading to maximizing the flight time of these UAVs.

### B. Cyber-physical systems

Cyber-physical systems (CPSs) are an innovative generation of systems with synergic cooperation between computational and physical potentials that can interact with humans through several new mechanisms [23]. Advancements in CPS should be developed to provide synchronized, distributed, rigorous, and responsive systems. The main success of CPSs in real-world applications is attributed to their reliability and predictability. CPSs cannot be deployed in real-world applications, such as traffic control, healthcare, and automotive safety, unless they are robust, reliable, and predictable. CPS design involves close integration of the cyber and physical worlds and consists of sensors that sense the real environments that generate huge data for analysis. The actuation information is then sent, such that the real world meets the desired outputs [24], as shown in Fig.3.

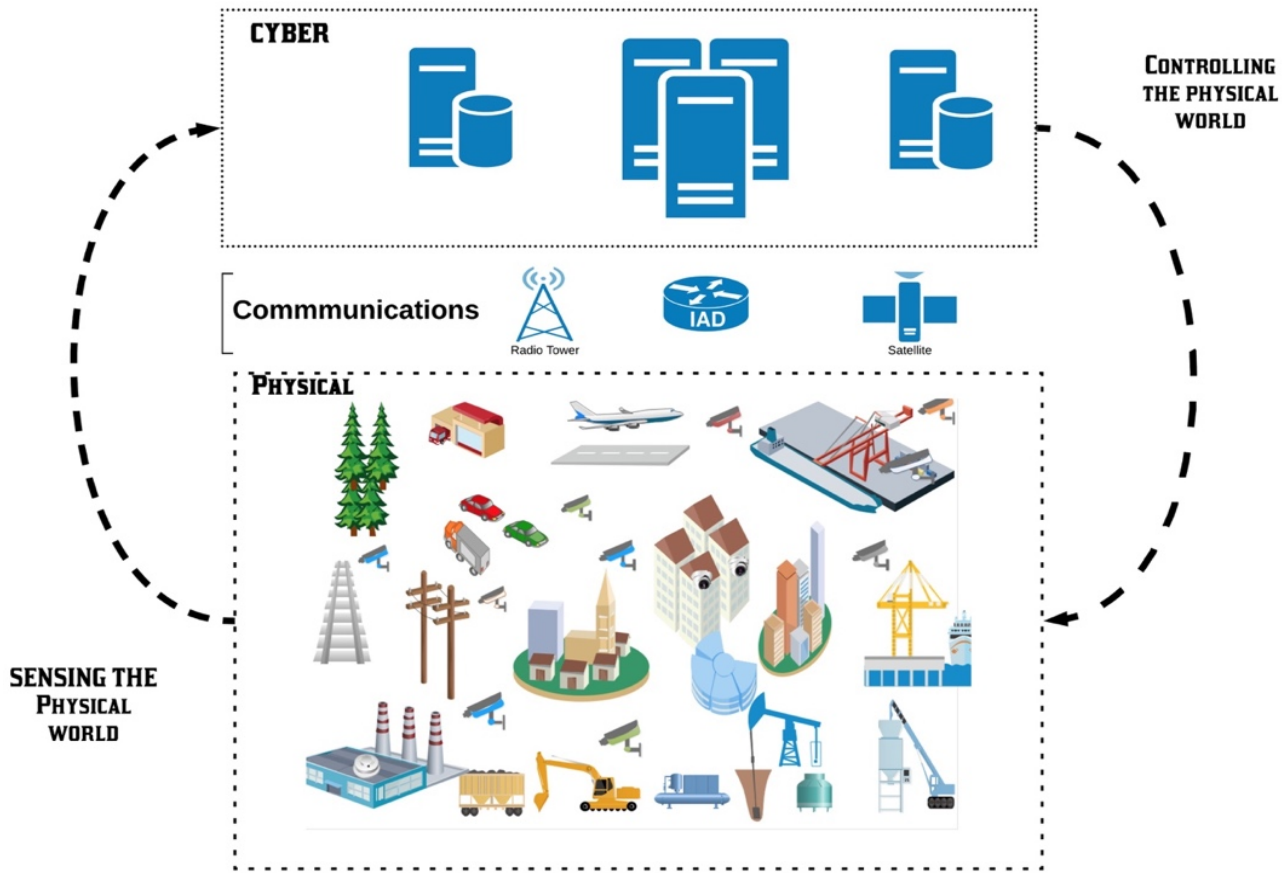

Fig.3. Cyber-physical systems

### C. Multi-UAVs in CPS applications

Several important CPS applications are discussed below. These applications will be used in the remainder of this survey to map and identify the design challenges of UAVs with the applications.

#### 1) Transportation applications

Intelligent transport systems represent essential part of realizing smart cities [25]. Modern transport systems are applications of CPS that aim to use information and communication infrastructure to provide connected and autonomous vehicles. However, according to the authors in [26], the full automation of vehicle systems requires reliable and efficient components to provide end-to-end transportation systems that can effectively and efficiently connect each part of the systems (i.e., vehicles and drivers, traffic police, and road emergency units). To meet such a requirement, UAVs can connect those components to provide intelligent, trustworthy, and automated vehicle systems [26].

#### 2) Construction and infrastructure inspection applications

Rapid growth in the use of virtual models in construction and infrastructure inspections offers a vital advantage by integrating virtual models with physical operations and control; this process involves real-time bi-directional coordination between virtual models and the physical construction [27]. The advantages of such integration are in the real-time inspections, quick decisions, and immediate collaboration between in-field operation and office teams [27].

UAV applications for monitoring construction, buildings, bridges, oil and gas pipelines, and power lines have grown. UAVs can be used in such CPS applications to inspect construction sites for providing safety to the worker, monitor works-in-progress, monitor oil and gas pipelines, and other construction and infrastructure inspections, specifically for hard-to-reach areas [28].

#### 3) Surveillance applications

The huge growth of CPS is pushed by many recent trends, such as the production of low-cost, embedded systems and efficient capability sensors and the recent abundance in Internet bandwidth, thereby resulting in several real-world applications. One current important application of CPS is in surveillance applications, including security surveillance, disaster surveillance, and other remote monitoring applications. Given the recent advances in image analysis methods such as deep learning algorithms, recent trends in CPS applications have been pushed toward developing situation-aware surveillance. Multi-UAVs can be integrated to CPS to provide flexible, high-mobility, and large-scale surveillance. They can also extend CPS surveillance into crowd surveillance and into areas that are unreachable and dangerous [29].

#### 4) Delivery of goods applications

Delivery of goods, particularly online products, is one growing application of CPS. Smart goods delivery can be achieved by using a radio frequency identification solution with mobile computing [30]. With the introduction of UAVs, the delivery systems of online goods are becoming increasingly practical, effective, and efficient [31, 32].

#### 5) Wireless & cellular systems applications

One rising application of UAVs is providing on-the-fly communications. UAVs can be used as base stations. The main advantages of using a UAV base station compared with a terrestrial base station are that the former has $3D$ (as it can move vertically), location-unrestricted, flexible, and mobile deployment compared with the 2D, permanent, and static deployment of the latter [17]. Therefore, considering the various advantages of UAVs in high-speed wireless communications, they are estimated to have an important role in future communication systems [33-35].

#### 6) Medical and healthcare system applications

CPSs are increasingly used in medical centers to improve the quality of healthcare services. CPSs are becoming a significant enabler for healthcare applications including in-hospital and in-home patient care [36].

Along with the advancement of CPSs, the development of capable drones also holds potential importance in revolutionizing the medical field and transforming the means by which healthcare is delivered. UAVs represent a promising technology that can result in vital advancements in clinical medicine [37]. They can be used to improve response times for out-of-hospital cardiac arrests, as well as enhance the time for vaccine, chemistry, hematology, and transfusion deliveries. UAVs can likewise be used to increase access to healthcare for patients who may not be able to have appropriate care because of distance or infrastructure restrictions [37].

### D. Related surveys and key contributions

Several surveys on UAV exist in literature. We discuss the existing surveys and provide differentiating aspects to contextualize the present survey within the existing literature. The study in [14] focuses merely on the communication perspective of UAVs. It provides a thorough review of quality of service requirements, connectivity, and data requirements, including minimum data to be transmitted over the network. Meanwhile, [14] covers the literature between 2000 and 2015 only. Although we also focus on the subject of UAV networking, we provide a more recent review on this subject and a thorough review on other subjects, such as UAV target coverage, path planning, and energy consumption. Similar to [14], [8] focuses on the networking side of UAV by comparing UAV networks, mobile ad-hoc networks, and vehicular ad hoc networks. Topics such as routing protocols and handovers are thoroughly investigated in [8]. In [38], a recent survey studies only the channel modeling of UAV communications.

The study in [17] provides a comprehensive tutorial on the potential use of UAVs in wireless networks. The authors investigate the use of UAVs in wireless networks with a focus on two main use cases of UAVs, namely, aerial base stations and cellular-connected users. The authors discussed and summarized the challenges, applications, and open issues related to each case. In similar research directions, [33] provides an overview of UAV-aided wireless communications with a focus on three use cases: UAV-aided ubiquitous coverage, UAV-aided relaying, and UAV-aided information dissemination. The study also discussed challenges and opportunities from the discussion of the considered cases.

The authors of [3] provide a comprehensive survey on UAVs and emphasize their potential use in the delivery of IoT services from the sky. They describe their envisioned UAV-based architecture and present the relevant key challenges and requirements.

A recent comprehensive survey [1] presents UAV civil applications and their related challenges. The authors review several civil applications in terms of the current state of the art and research trends, overall challenges, and potential future directions. The discussion mainly provides a review from a high-insight perspective; they presented several challenges of UAVs across different application themes.

Within the paradigm of WSNs, [39] presents a study on the coverage problem when the sensors are equipped with cameras for video surveillance applications but up to 2010 only. Although the study is not about UAV in particular, it provides a good review of coverage algorithms, node deployment, and coverage metrics. Similarly, [39, 40] provide a review of the coverage problem in WSNs but up to 2008 only. In particular, it focuses on finding the best solutions for the placement of sensors for optimum coverage. Furthermore, it presents different methods for data fusion for applications such as threat assessment. Another survey that focuses only on motion planning techniques for UAVs in different applications is presented in [41], but its scope is only up to 2010.

However, the aim of our survey is to provide a review that works in synergy with existing surveys to provide a comprehensive state of the art on this significant research topic. The key contributions of this survey are as follows.

- *A thorough discussion on the main challenges in designing multi-UAVs*. Five main challenges are discussed in this paper, namely, target coverage, path planning and image analysis, vision-based techniques, networking and cross-layer design, and flight control in designing multi-UAVs.
- *Presenting lessons learned*. The lessons learned for each design challenge are presented to emphasize high-order insights that emerge from the discussion and review of different approaches.
- *Qualitative and quantitative comparative studies*. We map the design challenges that are presented in each section to several vital CPS applications, such as transportation, infrastructure inspections, surveillance, delivery of goods, wireless & cellular systems, and medical and healthcare system applications. Comparisons and conclusions are then provided on the design challenges of UAV systems for each CPS application.

- *Presenting future research directions*. Potential research directions for multi-UAV systems are discussed based on the challenges discussed in this survey. These future directions are then mapped to several CPS applications to provide high-insight future directions for integrating UAVs into each application.

The rest of this survey paper is structured as follows. In Section II, we discuss the architecture and design challenges and issues for multi-UAV systems, which are also exhibited in Fig. 4, for a quick and conceptual glance. The figure highlights the overall picture, providing high level taxonomy of the challenges discussed in this survey. Sections III - VI discuss the major design challenges of multi-UAV cyber physical systems. Section VII discusses flight control methods. Section VIII proposes future and potential innovative ideas that could be investigated and studied. Finally, Section IX concludes this paper.

## II. ARCHITECTURE AND DESIGN CHALLENGES OF MULTI-UAV SYSTEMS

In this section, we depict the major architecture and design challenges and issues for multi-UAV systems, motivating and highlighting the trade-offs for distributed vs centralized processing and control of UAVs.

Single UAV testbed architectures usually focus on integration, and modification of the quadcopter modules to provide autonomous navigation and control. Multi-UAV systems focus, in addition to this, on achieving mission's goal collectively, through different drone formations that facilitate efficient coordination, leveraging centralized, distributed, or hybrid architectures. Hence, several research groups have recently developed UAV systems for education and civil applications purposes [42-45]. A single UAV testbed was developed by University of Colorado, Boulder for navigation in an unstructured GPS-denied environment, using vision-based navigation [44]. The ETH Flying Machine Arena [43] uses modified quadcopter platform to demonstrate several acrobatic and athletics control maneuvers. This architecture uses UAVs which were deployed in a 7 km2 area to test different operational conditions of the dynamic source routing protocol. The work in [42] shows the development of a low-cost UAV testbed using Goldberg Decathlon ARF model airplane. The work in [45] presents an approach for controlling a UAV through its front camera.

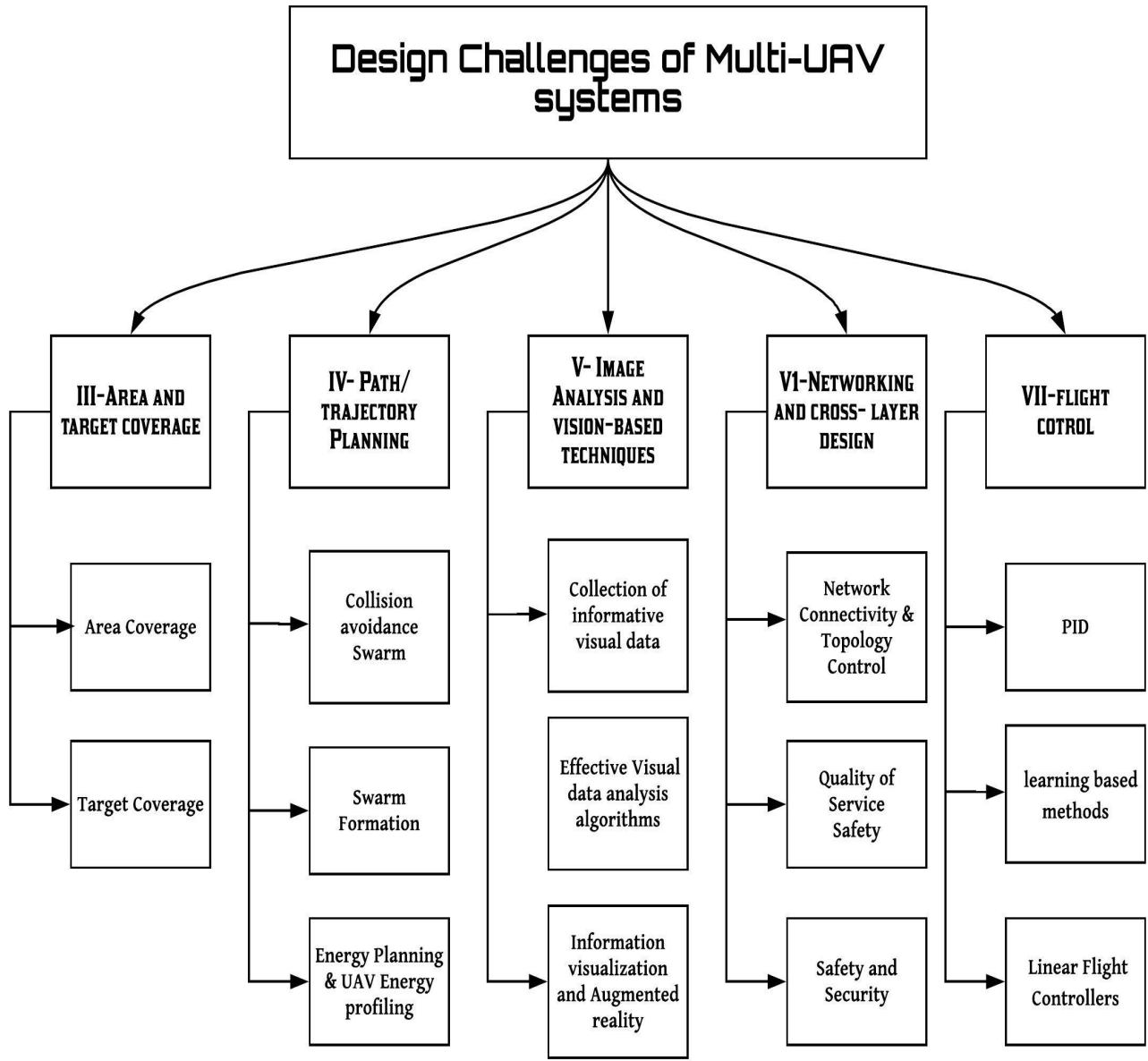

Fig.4. Design challenges of multi-UAV systems at a glance

For multi-UAV systems, leveraging centralized architectures for drone localization, navigation, and monitoring have the flexibility of providing efficient drone formations and accurate navigation through the global knowledge of the environment at the central controller. Such global knowledge, however, can only be facilitated through constant communication with all UAVs, deeming the system un-scalable for large number of UAVs covering large monitoring areas. In [46], the authors present the design, implementation, and evaluation of Up & Away (UnA) a generic two-tier testbed developed for UAV related CPS experiments. UnA aims to use a top tier of centralized camera network for localization, and a low tier drone network for surveillance and monitoring, while using centralized control server for both tiers. However, the UnA testbed has several limitations. The cheap off-the-shelf AR.Drones used in UnA possess limited built-in sensors (e.g., inaccurate compass and altitude sensors) causing serious problems in controlling the rotation and altitude for accurate navigation. Furthermore, errors in drone placement and maneuvering inherently appear due to the simplicity of the control algorithms they have adopted. Fig. 5 gives a general concept of view on two-tier testbed.

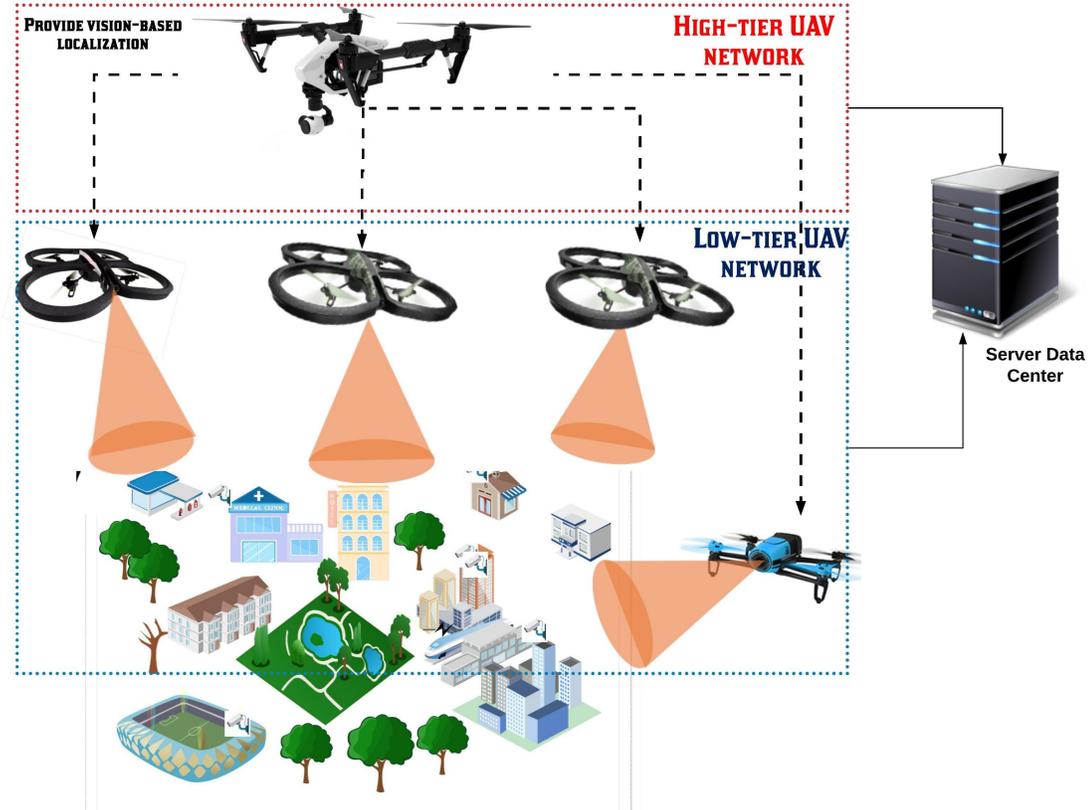

Fig. 5. Two tier framework of the multi-UAV system.

It was then learned in [47] that leveraging edge computing [48] through pushing computations partially toward UAVs while maintaining high-level picture at the central controller will potentially enhance scalability through hybrid architecture that leverages distributed UAV-based monitoring & autonomous UAV navigation, while maintaining a centralized high-level light-weight navigation and control. Authors in [47] introduce Drone-Be-Gone (DbeG) as an inexpensive, decentralized and accessible CPS testbed using off-the-shelf UAV with an external processing unit (EPU). DbeG's architecture has 2 modules - the Central Node and the Client Node. The Central Node acts as indoor-navigator, providing vision-based localization and high-level navigation commands to individual UAVs. The Client Node comprises of an AR.Drone quadcopter and an EPU allowing low level autonomous navigation and local processing. Such autonomous navigation and distributed processing at the UAV level enhance the scalability to cover large areas. However, distributed navigation and processing poses the challenge of individual UAVs losing the connection with the Central Node, potentially causing safety and connectivity issues. A comprehensive hardware integration study has also been conducted to assess the feasibility of two different EPU platforms, namely Raspberry Pi and Intel Edison. In addition, an inexpensive inertial measurement unit is attached to the EPU to resolve AR.Drone's yaw angular errors due to magnetic distortions. Thus, DbeG harbors prime qualities as a CPS testbed such as inexpensiveness, flexibility, controllability of multiple UAVs and localization.

While decentralized architectures provide the ultimate scalability for large area coverage through leveraging distributed UAV navigation and control, they pose several challenges in terms of maintaining connectivity and providing accurate coverage simultaneously, which might have implications on safety and robustness of such systems. The GRASP testbed [49] uses off-the-shelf high-end quadcopter to demonstrate multi-robot control algorithms. Authors discuss the autonomous navigation and control of individual UAV, and formulate the aerodynamic control of the UAV to address the issues associated with distributed processing, including collision avoidance, and aerodynamic interaction amongst UAVs due to rotor downwash, which influence UAVs trajectory, hence robustness of the entire system. Moreover, for the application of searching and tracking of moving convoys over urban environments using UAVs, the authors in [50] designed a control logic to test whether multi-UAVs can autonomously cooperate and succeed in switching between different operation modes, which include takeoff mode, flying to area of operation mode, searching mode and tracking mode.

Assuming that UAVs know the extracted road map information for the area of interest, the authors use their developed *3D* hybrid multi-UAV simulation system to test their control logic in a $2km \times 2km$ area with 23 roads distributed through the map. They developed an autonomous takeoff strategy as well as a search algorithm that combines a road map and a probability map information. Their results show that decentralized UAV control is efficient and autonomous UAV operation is feasible.

In the following, the most vital and important challenges of implementing multi-UAV systems are discussed in the order given in Fig. 4. First, each challenge is explained, then the relating threads are opened followed by examples, their difficulties and advantages.

## III. AREA AND TARGET COVERAGE

Literally, camera coverage is the amount of footage shot and different camera angles used to capture a scene. This definition, however, applies to surveillance applications but coverage more professionally is referred to as the state when there is at least one camera viewing the target with acceptable quality. Technically, the area monitored by cameras is called Field of View (FoV) or Angle of view (AoV) which is within [0; 360°] degrees. The triangular area in Fig. 6 depicts the FoV of a camera sensor node, where is called the FoV vertex angle, is defined as the camera orientation and $R_{min}$ and $R_{max}$ are the minimum and the maximum sensing ranges of the camera, respectively. Any targets could be monitored only if its underlying distance to the camera lies within the camera sensing range $R_{min} \le r \le R_{max}$ . Although the scenario of having multiple cameras viewing the target may not be practical for some applications, such assumption has been used in many coverage studies [51-55]. Other applications, however, require full coverage of a target by one UAV [56]. That is, a target is fully covered when it is seen fully by one camera from an angle. Target coverage maximization is generally NP-complete for most variations, e.g., [39] and [57]. However, several studies address this intractability through approximations and assumptions. The problem of coverage in monitoring applications is majorly divided into two categories with their sub-categories that are elaborated below and in Fig. 7 concisely.

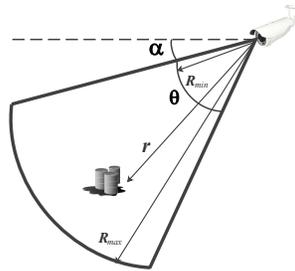

Fig. 6. Concept of coverage at a general glance.

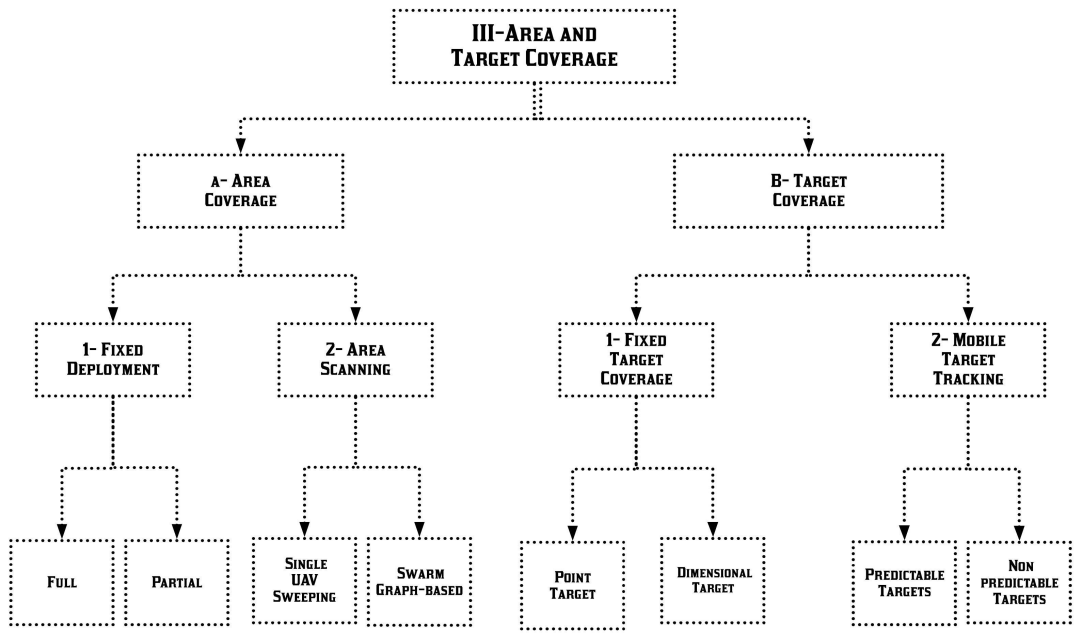

Fig.7. Concept of coverage at a general glance.

## A. Area Coverage

The main objective here is to cover the entire area regardless of objects/targets inside, which is usually practical when surveillance is conducted in a controlled or reasonably small (dense) area. Another slight variation of such objective is to cover the targets with known or partially known locations in a controlled area. Depending on the number and mobility of the cameras, the objective in this case may be to either minimize the number of fixed cameras deployed to cover certain area, or to minimize the time required to cover the area using a number of mobile cameras to detect any events within. Therefore, we classify the work in this area into two categories, namely: Fixed camera deployment and Area scanning.

### 1) Fixed deployment

In this category, the objective is to mainly minimize the number of fixed cameras deployed to cover the area and maximize the probability of detection of any event that might happen. Worth noting here that fixed camera deployment may include cameras with pan feature i.e., variable orientation. Many of the solutions in this category are based on the traditional Art Gallery problem [58]. The objective in this problem is to find the location of the minimum number of guards in an art gallery (i.e., controlled area), such that every point in the area is seen by at least one guard. Coverage could be partial or full based on the application importance. Sometimes, there is a need to observe a target from any possible angle and in contrast, there may be targets that only a partial view of some parts of them will suffice, depending on the application requirements. Continuous-space Art Gallery problem is deemed both NP-Complete and APX-hard [59, 60]. Therefore, for polynomial-time log-approximation solutions, researchers sometimes use space discretization by considering only potential points or pixels for positions and/or poses of the cameras [61, 62]. The angular view or field of view (FoV) and limited range for each guard, however, makes the area coverage problem with cameras more complex. Even after discretization of space and camera poses, potential solutions remain un-scalable [63]. Therefore, a number of heuristics using angular sweeping for camera pans have been proposed, which proved to have performance close to optimal, although still require high computational complexity. Angular sweeping is also leveraged for a set of pre-deployed cameras to minimize the subset of such cameras which are to be activated to attain maximal coverage [52]. Authors propose both an integer linear programming formulation and a sub-optimal greedy-based approach to maximize the number of covered targets, while minimize the subset of cameras activated within a randomly deployed camera.

### 2) Area scanning

In this category, the objective is to minimize the number of mobile cameras and the time required to scan the area under surveillance, while minimize the probability of missing an event. In the robotics community, the area coverage becomes how to move the robot swarm and direct sensors to optimize a coverage objective. Applications in this category such as mine sweeping and lawn mowing may require a single mobile robot or UAV to scan the area in a minimum time. In [64], authors conduct area scanning through subdividing the work-area into disjoint cells corresponding to the square-shaped tool used for sweeping. Then, they follow a spanning tree of the graph induced by the cells, while covering every point precisely once to minimize the coverage time. Authors in [65-69] have designed distributed gradient-based optimization methods to control the robot swarm to maximize the detection probability of any event in the area of interest.

UAVs open the door for many other applications such as forest fire monitoring, border patrol, aerial mapping, farming, in addition to military applications. [70] generates waypoints for individual UAVs such that the visited area within a bounded region is maximized during a fixed interval of time. First, they segment the work-area into cells, and create a graph with all cells represented as vertices, while connections between cells is represented using edges. They generate ground paths by constructing a tree consisting of a set of vertices and edges for each UAV assigned to the task, and then the trees are connected together such that a single tour about the resulting spanning tree will survey the entire region. Finally, the resulting ground path is transformed into a series of aerial waypoints for the UAV that results in the camera image following the previously calculated ground path. In [71], authors address the objective of minimum time coverage using a group of UAVs equipped with image sensors. There too, they model the area as a graph whose vertices are geographic coordinates determined in such a way that a single UAV would cover the area. The minimum coverage time solution is then achieved using a mixed integer linear programming problem, formulated according to the graph variables to route the team of UAVs over the area. The problem of swarm distribution for surveillance of a controlled area with possible physical obstacles and/or no-fly zones has been addressed in [72]. Authors adopted an evolutionary-based approach using Particle Swarm Optimization (PSO) to calculate the 3D trajectories that UAVs use to move from an initial position (the depot) to the final location to cover the areas of interest. To simplify the localization of the UAVs, the relative distance between the UAVs was constrained as part of the PSO optimization to facilitate the evolutionary movements.

## B. Target Coverage

This is defined as finding the minimal number of cameras and their optimal camera placements and/or orientations using any directional sensor so that each target is fully or partially covered by at least one camera. Another definition is that given a fixed number of cameras, what is the maximum number of targets so that each target is fully or partially covered by at least one camera. Based on the fact that the target and the camera are moving or fixed entities, there are four possible situations to tackle as follows:

- The camera and the target are both fixed
- The camera is mobile while the target is fixed
- The camera is fixed but the target is mobile
- Both camera and target are moving

The taxonomy above gives versatility to the problem of target coverage, and its proposed solutions. We summarize the first two categories in the following section (III-1), while we cover the last two categories based on mobile targets in section (III-2)

### 1) Fixed Target Coverage

Solutions in this category can be categorized based on how the fixed target is modeled, namely: point (dimensionless) target, or dimensional target.

#### a) Point (dimensionless) target

The optimal solutions to these variants are shown to also be NP-complete, and therefore the authors proposed heuristics. Within this class, the camera locations are assumed to be known and fixed or chosen from a pool of formerly selected random positions. First, in the work presented in [51-53, 55] several solutions have been proposed that are based on simplifications. A heuristic solution is proposed in [51] in which it was assumed that the camera location are fixed along with discrete camera pans. In [53], a subset of randomly scattered directional sensors is selected one at a time. A similar work is presented in [53] where an active set of camera sensors and their directions are selected from a larger pool of cameras that were placed beforehand. Rotating directional sensors are utilized in [55], to maximize various coverage objectives. The studies in [73-77] consider fixed deployment of anisotropic sensors to cover a set of point targets. They map the variations of such problems into the famous Art Gallery problem, and use algorithms based on integer linear programming to find the best sensor topologies to cover the targets.

Several algorithms targeting low-complexity solutions have been proposed in [78-80]. The problem of low complexity placement and coordination of an unknown number of mobile cameras to cover arbitrary set of targets are considered therein. This problem is addressed as an unsupervised clustering task. A set of proximal targets are clustered together, whereas the camera location/direction for each cluster are estimated individually. In [80], Cluster-First (CF) iterative method is implemented as well as Simple Cover-Set Coverage Method (SCSC) to find the optimal camera position and orientation to cover one cluster. In each iteration, the targets are divided into a number of clusters, $k$, as described. Then the proposed SCSC algorithm is used to find the location/direction of the single camera that is to cover all the targets in each cluster. Although this task may be inconceivable, ultimately, the number of targets that have been covered are calculated at the end of these steps. In the case that a target is left uncovered, the number of clusters, $k$, is increased either using binary search, a multiplicative increase/decrease scheme or uniform steps. Then a reiteration is performed until no targets are left uncovered. They state their problem as finding the radial sector, with minimum possible radius $r$, its location and direction to contain an ellipse originated at $(0; 0)$ and parameters $a$ and $b$. Afterwards, they place the center of a circular sector of angle on the major and minor axes of the smallest ellipse (in the area) containing all the targets in the target set referred to as opt, such that the upper and lower boundaries are its tangent. If the resulting location of the sector vertex is within 0 and $R_{max}$ to all targets, this sector can cover them all. They further show that the maximum distance between the camera and the farthest possible target, $d_{max}^1$ and $d_{max}^2$ are obtained as follows:

$$d_{max}^1 = a + \sqrt{a^2 + \frac{b_2}{\tan^2(\theta/2)}}$$

$$(1)$$

$$d_{max}^2 = a + \sqrt{b^2 + \frac{a_2}{\tan^2(\theta/2)}}$$

[78] compares two clustering-based algorithms, k-camera clustering and cluster-first algorithms, to provide centralized computationally efficient heuristics. The required number of cameras approach those obtained via near-optimal methods as the camera's coverage range, angles of view, or the number of targets increase. In [78], the efficiency of the converted-feature clustering in the same context is studied. They show that the complexity of $k$-camera algorithm could vary based on the method to increase the number of cameras. The unit increment in their analysis result in $O((K - k_0)NM$ where $k_0$ and $K$ are the initial and final number of cameras.

In [79], the authors proposed to implement two clustering algorithms: Smart Start $K$-Camera Clustering (SSKCAM) and fuzzy coverage algorithm. SSKCAM starts from a given set of clusters identified by an off-the-shelf clustering algorithm and iteratively adjust clusters, based on the coverage status of their comprising targets. Then, it recalculates the camera location/direction for each individual cluster until convergence is achieved. Fuzzy Coverage algorithm builds overlapping (or fuzzy) clusters. As a result, a target can belong to multiple clusters.

As far as the concept of fixed target coverage is concerned, there should be some defined metrics to evaluate the performance of different proposed algorithms for multi-UAV systems to cover fixed targets. In the literature, there are three common metrics reported, which are: 1) Fraction of uncovered targets, 2) Execution time, and 3) Number of cameras. The behavior of each of which will be elaborated and compared versus three other varying parameters 1) Number of Targets, 2) FoV, and 3) $R_{max}$.

The fraction of uncovered targets shows, for fixed number of cameras, the ratio between the number of targets with no camera pointing at them, and the total number of targets. This in fact could be a powerful metric to refer when evaluating the performance of an algorithm. Execution time, captures the complexity level of an algorithm in terms of both implementation and time delay that an algorithm of interest imposes for a set of targets to be covered. The number of cameras on the other hand, could well demonstrate the ability of a technique to cover fixed targets since it is expected to be far less than the number of targets. Therefore, the lesser the number of cameras, the better the performance of an algorithm when it comes to target coverage problem.

In addition, [80] proposes a computationally light-weight heuristic called cluster first (CF), where the number of used mobile cameras is close to those found by near-optimal algorithms. Specifically, they address this problem for non-uniform target distributions that naturally form clusters. They also apply two specific clustering algorithms as representatives of centroid and non-centroid based clustering algorithms namely $k$-means, and kd-tree [81], and denote them by CF-$k$-means and CF-tree.

We conducted a comprehensive performance comparison of the algorithms proposed in [79, 80], along with greedy and dual sampling (D-Smp) algorithms in terms of uncovered fraction, execution time and number of cameras when the number of targets runs from 0 to 200 in Fig. 8 . It is shown by Fig. 8 (a) that the greedy and D-Smp algorithms achieve perfect coverage (in the tested scenarios), while SSKCam and fuzzy coverage leave a fraction of targets uncovered. This is because the authors in [79] allow their solutions to find imperfect coverage at the cost of getting lower computational complexity. SSKCAM demonstrates a better coverage performance compared to fuzzy coverage, however, leaving a fewer number of targets uncovered. It could also be noticed that the CF algorithm requires a number of cameras that is very close to that required by Greedy and Dual-Smp. In terms of complexity, Fig. 8 (b) shows that fuzzy coverage has a very low execution time compared to the other algorithms. Also, SSKCAM does better than D-Smp and exhibits better scalability as the number of targets increases. For fuzzy coverage, the advantage in computational complexity is a result of the onetime operation of clustering (instead of iterating) and camera. configuration computation. However, this leaves more targets uncovered in comparison to SSKCAM. As can be seen in Fig. 8 (c), permitting compromise in fraction of covered targets also yields camera numbers obtained to be very similar to those from greedy and D-Smp, for SSKCAM, and even lower (better) for fuzzy coverage.

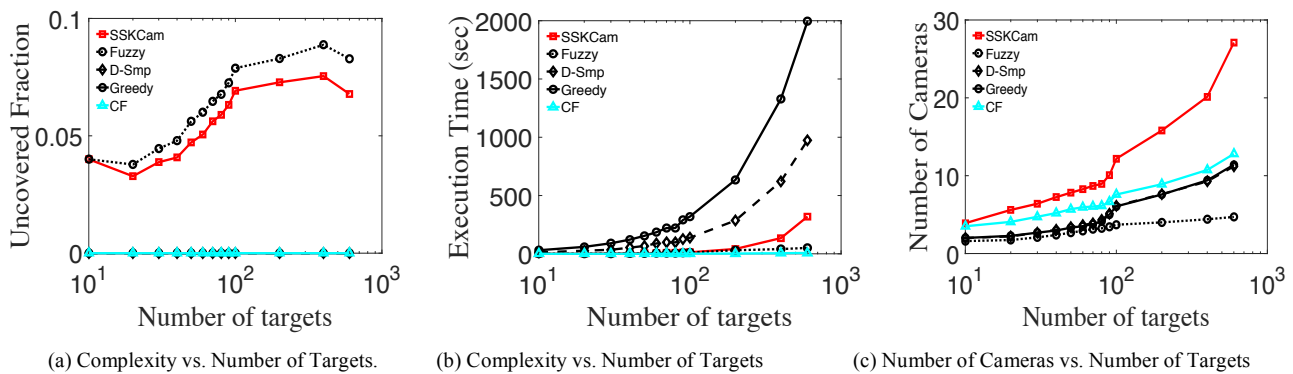

(a) Complexity vs. Number of Targets.  (b) Complexity vs. Number of Targets  (c) Number of Cameras vs. Number of Targets

Fig. 8. Performance comparison of different algorithms in terms of uncovered fraction, execution time and number of cameras versus Number of Targets.

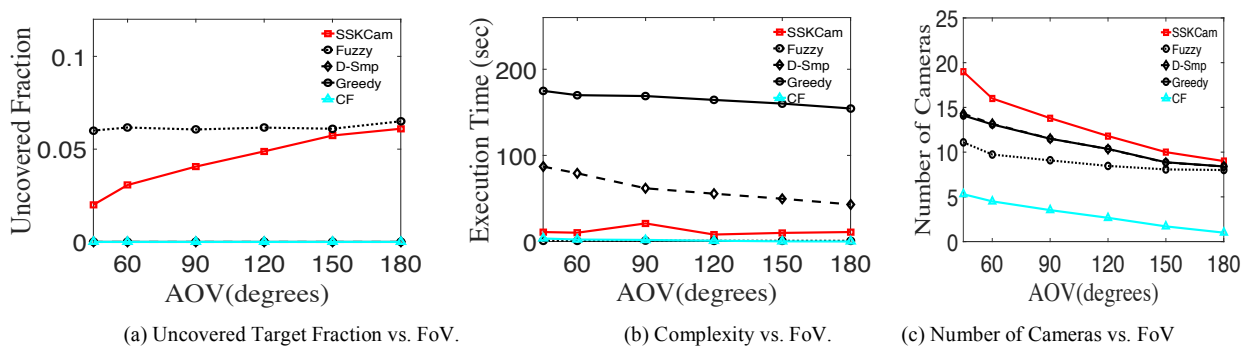

(a) Uncovered Target Fraction vs. FoV.  (b) Complexity vs. FoV.  (c) Number of Cameras vs. FoV

Fig. 9. Performance comparison of different algorithms in terms of uncovered fraction, execution time and number of cameras with varying FoV.

Note that the latter advantage for fuzzy coverage comes at the price of lower coverage fraction.

The performance of all algorithms for different values of FoV: 45º (FoV of an un-zoomed webcam), 60° (FoV of vertical camera of parrot AR.Drone 2:0), 90° (close to the FoV of frontal camera of Parrot AR Drone 2:0), 120°, 150°, and 180° for fish-eyed lenses are compared to each other in Fig. 9. It can be noticed from these figures that, the number of required cameras for all algorithms decreases (and almost converges) for wider values of FoV. Similar effect happens to them in terms of complexity. But in contrast, the fraction of uncovered area for two algorithms Greedy and D-Smp, outweigh that demonstrated by SSKCAM and fuzzy coverage.

As illustrated in Fig. 10, The number of required cameras decreases for all algorithms as $R_{max}$ increases. This quantity was also quite similar across all the algorithms. Based on the results depicted in Fig. 10 (b), the execution time of the fuzzy coverage and SSKCAM decrease as $R_{max}$ increases, D-Smp exhibits an exponential increase in execution time moreover. In Table 2 , we summarize the comparison outcomes for all algorithms.

### b) Dimensional target

Most of the solutions above modeled the target as a point (i.e., dimension-less) or they assume the target is fully covered if only the center of the target is within the FoV of at least one camera, to simplify the coverage problem. However, flying on low altitudes, UAVs can potentially monitor fine-grained target details, and hence, [82] is one of the early research attempts that started to look into realistic target models as line segments ($1D$), or as $2D$ dimensional targets. Note that, by the term $1D$ we mean that one dimension of the object is far bigger that the other. Imagine the case that the area of interest is a power station and the target is to cover only the main pipeline feeding fuel into the boiler. The designer here is confronted to the problem of covering a one-dimensional target (modeling the pipeline as a line segment, neglecting the radius), which is fixed. Thus, dividing the pipeline into multiple parts will convert the problem into fixed and multi-target coverage, which could solve the problem readily, using many of the approaches discussed above. In other words, cameras can be situated at fixed points that each is able to cover one part fully.

For many real-world applications however, the $2D$ target view and coverage is needed. Consider the case where UAVs (i.e., mobile cameras) are sent for investigation of an infrastructure (e.g., pipeline grid, or a bridge, etc.) from the top view (see Fig. 12), because the deployment of infrastructure sensors at or on the target is environmentally or strictly prohibitive. In this case, dividing the infrastructure into $2D$ targets could cause the number of the targets to grow large, making the issue of coverage and tracking twisted towards how to manage and control a large number of UAVs for efficient coverage, in which case, collision avoidance and connectivity. issues between the UAVs at any instance should be tackled. Hence, there are still too many issues that are caused by switching from $1D$ coverage and tracking problem to $2D$ applications, let alone adding another dimension on top of them and forming $3D$ coverage and tracking problem. The slightly good news is that, not all the aspects of one $3D$ target are important at all times. There exist some objects that although are categorized under $3D$ target coverage problem, we can model such target as a set of $2D$ targets, each with a direction. In other words, only the directional view of some critical sections of such targets suffices. Fig. 13 demonstrates how dividing the infrastructure of a bridge into a set of critical areas could facilitate the problem of $3D$ coverage into $2D$ coverage problem that we shall call this type of target coverage as directional coverage [82].

Even then, another challenging issue that might arise is occlusion, where targets can block the camera view from each other, making target coverage solutions more intractable [83]. Figure 14 shows the scenario of a $3D$ target divided into $2D$ directional targets where target $T1$ blocks the camera view from target $T2$ .

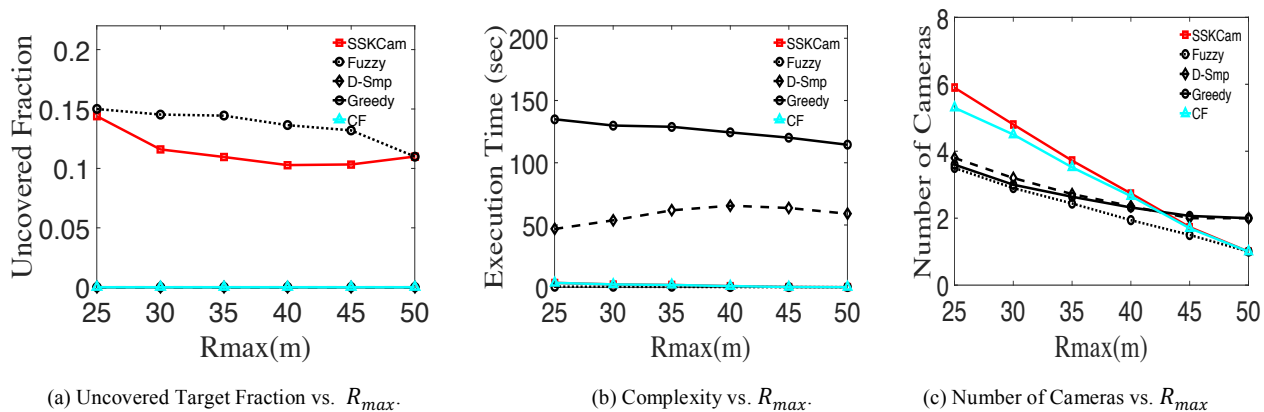

(a) Uncovered Target Fraction vs. $R_{max}$.

(b) Complexity vs. $R_{max}$.

(c) Number of Cameras vs. $R_{max}$

Fig.10 Performance comparison of different algorithms in terms of uncovered fraction, execution time and number of cameras versus Number of Targets when $R_{max}$ varies.

TABLE 2
COMPARISON BETWEEN FIXED-TARGET COVERAGE ALGORITHMS

| | SSKCAM | Fuzzy | D-Samp | Greedy | CF |
|---|---|---|---|---|---|
| **Target Density** | | | | | |
| **Uncovered Fraction** | 6% (average) | 7% (average) | Perfect Coverage | Perfect Coverage | Perfect Coverage |
| **Execution Time** | Middle | Lowest | High | Extremely High | Low |
| **Number of Cameras** | Middle | Lowest | Highest | Highest | Highest |
| **FoV** | | | | | |
| **Uncovered Fraction** | 4% (average) | 6% (average) | Perfect Coverage | Perfect Coverage | N/A |
| **Execution Time** | Middle | Lowest | High | Extremely High | Low |
| **Number of Cameras** | Highest | Lowest | Middle | Middle | Lowest |
| **$R_{max}$** | | | | | |
| **Uncovered Fraction** | 11% (average) | 13% (average) | Perfect Coverage | Perfect Coverage | N/A |
| **Execution Time** | Middle | Lowest | High | Extremely High | Low |
| **Number of Cameras** | slightly higher | lowest | slightly higher | slightly higher | slightly higher |

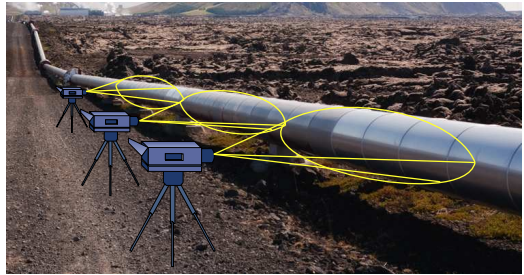

Fig.11.1$D$ target coverage illustration

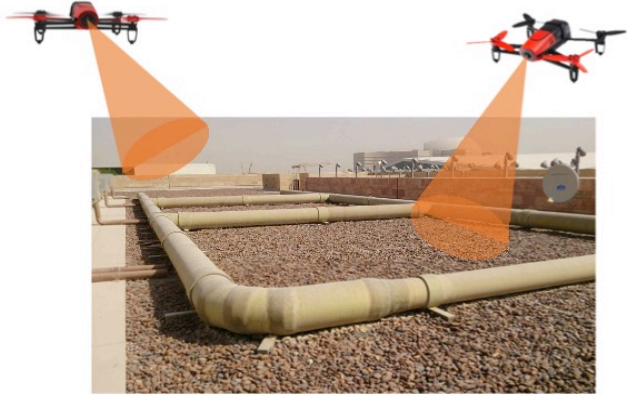

Fig.12. A picture of a 2$D$ pipeline and UAVs flying on top

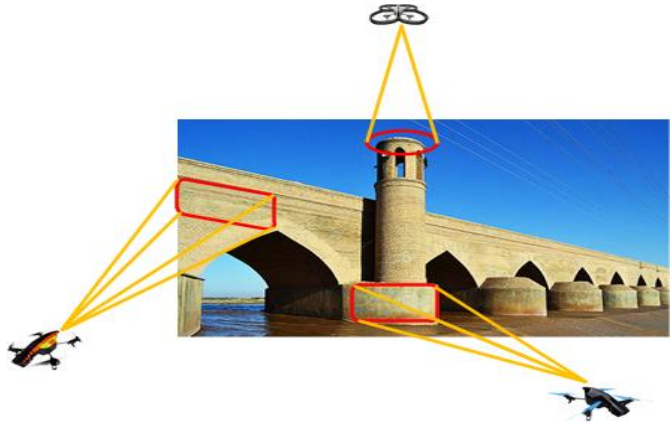

Fig.13. Dividing infrastructure resources into 2$D$ oriented targets.

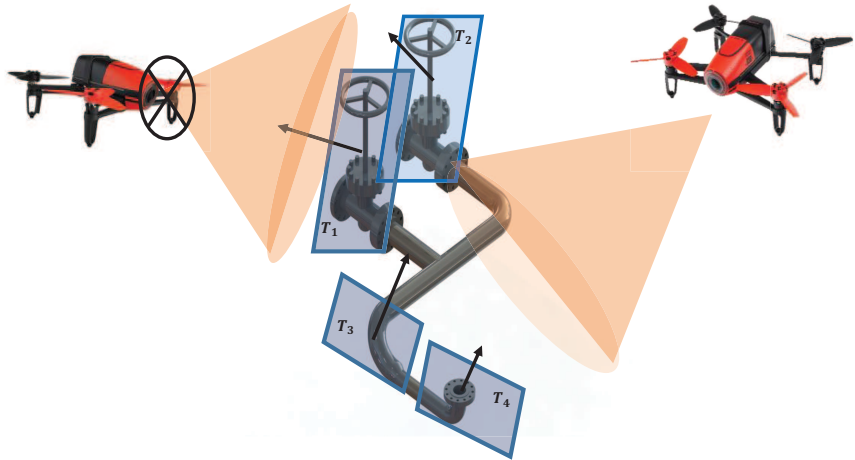

Fig.14.3*D* target coverage with occlusion

### 2) Mobile Target Tracking

Once the target is mobile and the camera has got the ability of moving (which UAVs have this option definitely), the problem of target tracking emerges. There should be some algorithms to be proposed to enable cameras to track a moving target wherever it relocates. One potential category of tracking solutions may neglect the temporal dependency of target locations throughout the time, and tackle the target tracking problem as independent coverage problems at every time instance. In a controlled area, this assumption might hold realistic, and with limited camera movements (i.e., variable pan/tilt) the targets can be covered at all time instances [54]. However, depending on the target mobility pattern and the size of the surveillance area, such solutions may be deemed inefficient in terms of continuous UAV control and navigation, and in many cases may require large mechanical energy to continuously navigate the UAVs and keep the targets covered at all times. Therefore, we classify the behavior of target mobility into two main classes namely predictable and non-predictable. The advantage of predictable targets is that by knowing the nature of their movements, one could optimize the camera movements not only for current UAV location, but also for future predicted locations.

Target tracking has also been a natural topic of great interest in CPS applications, which remains challenging because of several factors including illumination variation, occlusion, viewpoint variation as well as background clutters [84].One promising solution is the popular tracking-by-detection discriminative approaches [85], where numerous training and learning methods have been developed to further improve the classifiers [86, 87]. Such visual tracking and image matching methods can also be employed for vision-aided localization of UAVs [88]. In addition, a prediction model for the fisher information matrix is developed in [89] for target tracking of moving RF sources.

#### a) Unpredictable targets

In support of the above explanations and for unpredictable targets, a single UAV target tracking problem has been addressed in [90]. Authors propose two optimization-based control strategies to minimize a UAV expected cumulative cost function over a finite time horizon for tracking a single target. The first approach is game-theoretic based on two-player, zero-sum game with perfect state feedback and simultaneous play. The second is a stochastic optimal control approach that leverages the framework of Markov Decision Processes (MDPs) to estimate the state probabilities for target mobility. [91] studies optimal coordination amongst multiple UAVs to perform vision-based target tracking of a single unpredictable ground target. They leverage dynamic programming to devise an optimal control policy that minimizes the expected value of the fused geolocation error covariance over time.

Tracking multiple unpredictable ground targets with multiple UAVs has been considered in [92, 93]. In [92], they utilize a two-phase approach as a suboptimal solution of target clustering/assignment and cooperative standoff group tracking with online local re-planning. Within each target cluster, a nominal orbit radius of each UAV is set to ensure that each UAV maintains visibility of all targets in the corresponding cluster, while maintaining enough accuracy of estimating (monitoring) each target in the cluster. The UAV groups follow a set of rules to discard stray target vehicles as well as ensure successful target handoff between teams (clusters). [93] treats the problem of tracking multiple targets with multiple UAVs as a partially observable Markov decision processes (POMDPs), wherein heuristics are employed in the approximate solution to overcome the limitations of short planning horizons in the presence of occlusions. They employ new approximation method called nominal belief-state optimization (NBO), combined with other application-specific approximations to produce a practical design that coordinates the UAVs to achieve good long-term mean-squared-error tracking performance in the presence of occlusions and dynamic constraints.

### b) Predictable targets

Predictable targets refer to the scenario where future target locations are deterministically known, or can be approximately estimated using statistical models. Optimizing the coverage for target tracking using some mobility prediction techniques will be crucial for practicality. Such mobility prediction techniques facilitate future coverage of moving targets within a significant time range. The assumption stands by knowing how the targets move partially, with the help of some mobility models that best approximate the target movements, such as Manhattan Grid (MG) [94], Reference Point Group (RPG) [95] and Random WayPoint (RWP) [96, 97], amongst others [98, 99]. Manhattan mobility is particularly feasible in sites where personnel and vehicles have certain roads and finite paths to move from one place to another. Random mobility models could also be used to approximate the movements of targets, e.g., objects or fluids in free space.

To address the problem of target tracking using mobility predictions, one approach can be as simple as generating virtual targets that represent the location of the targets in future time instances before running the coverage algorithms. Such simple idea allows us to cover the targets for the current and future times, while trying to minimize the UAV locations to monitor these targets. Indeed, the more accurate the prediction model is compared to the actual mobility of the moving target, the smaller number of UAV locations we can have to monitor such targets.

Aside from this, one rigorous challenge with target tracking is accuracy of the UAVs following the target, i.e., the UAVs should be so adaptable and flexible to take actions when needed. Since the drone is a mechanical object flown into the air, when the target is quite agile, it would be impossible or at least hard to have a drone possessing the capability of being as quick as the target. This gives rise to the issue of implementing versatile and light weight drones, while emphasizes the practicality of target mobility predictions.

Regarding this class of tracking multiple predictable targets, [100] proposes computationally efficient approaches that address the problem of mobile target coverage practically. This is done by clustering targets, while calculating the camera locations and poses for each cluster independently using a cover-set coverage method. Three computationally efficient approaches are developed, namely: Predictive Fuzzy Algorithm (PFA), Predictive Incremental Fuzzy Algorithms (PIFA) and Local Incremental Fuzzy Algorithm (LIFA). The objective is to find a compromise between coverage efficiency, traveled distance, number of drones used for tracking and complexity.

The targets move according to mobility patterns, including RWP, MG, or RPG as explained above. In order to divide targets into clusters, Fuzzy C-Means (FCM) algorithm [101] is used. The FCM algorithm partitions a set of targets, $x_i$, into a set of C fuzzy clusters. The FCM algorithm returns a partition matrix with the degree to which element $x_i$ belongs to cluster $c_k$. In other words, any target has a set of coefficients giving the degree of being in the $k th$ cluster.

PFA leverages the FCM algorithm defined above to divide the targets into clusters, allowing overlapping (fuzzy) clusters. It predicts targets' movement and includes the predicted targets' locations as new (i.e., virtual) targets before dividing the targets into clusters. After x given time steps, the FCM algorithm is run again and new clusters and the camera locations and orientations are determined. The normalized complexity overtime of the PFA was shown to be $O(f(\bar{N}/\nabla t)$ where $\bar{N}$ is the total number of current and future target locations at timestep $t \rightarrow t + \Delta t$ and $f(\hat{N}) = M\hat{N}2\pi/\Delta\phi + MC\hat{N}$ , $C$ is the number of clusters (i.e., used cameras), $\Delta\phi$ is the angular sweeping step for running the cover-set [79] , and $M$ is the complexity of the search scheme.

Similarly, PIFA uses the Fuzzy Coverage (FC) algorithm and predictions based on statistical knowledge of the targets' movement. However, instead of finding the best location and orientation of the camera for all predicted targets' locations, it gives priority to the closest prediction in terms of time without impacting the coverage ratio found in the previous step. After x timesteps, the FC algorithm is run again and new clusters, camera's location and orientation are determined. The normalized complexity of PIFA is $O(f(N)/\Delta t + N2\pi/\Delta\phi)$ .

Notice here that for high target mobility $N \ll \bar{N}$, which allows PIFA to be less complex compared to PFA, with the cost of possibly missing targets. LIFA is also based on the Fuzzy Coverage (FC) algorithm. However, in contrary to PIFA, it does not use predictions. The targets are first divided into fuzzy clusters and the camera position and direction are determined to cover the clusters. Keeping the targets associated to the same cluster, only new camera's location and orientation are computed at each timestep. The normalized complexity of LIFA is $O(f(N)/t_{max} + N2\pi/\nabla\emptyset)$ as FCM is only run once at the start. Hence, LIFA is expected to provide low complexity solution for application scenarios where targets tend to move in well-behaved clusters (clusters do not change over time, e.g., group mobility pattern).

The above algorithms have been examined, considering the coverage ratio (the percentage of targets covered by at least one camera), number of mobile cameras required for tracking, total traveled distance by mobile cameras, and time complexity of the coverage algorithms.

The coverage ratio with respect to the number of targets is shown in Fig.15. First, all algorithms can maintain the coverage ratio above 60%, for all mobility patterns. As expected, PFA maintains the highest coverage ratio, followed by PIFA and LIFA. These differences are due to the intrinsic mechanisms of each algorithm. LIFA does not rely on future locations of the targets and does not update the clustering to re-position the mobile cameras, while PFA and PIFA use prediction and adapt the number of drones accordingly. Worth noting here that the favorable results obtained in coverage ratio by PFA has a cost in terms of the number of drones required, traveled distance and time complexity.

The number of drones vs. the number of targets is shown in Figures 16(a), 16(b) and 16(c) considering the RWP, MG and RPG mobility models. LIFA, by design, is very efficient in terms of the number of drones needed and these results are similar to those obtained in [79]. LIFA does not consider prediction and hence, does not need to cover all or a subset of the predicted target locations. It does not update the number of cameras in the field during the simulation. As a result, a lower number of mobile cameras is required and this number remains constant. As observed, the MG model slightly increases the number of cameras needed for PFA and PIFA when the RPG model reduces drastically the number of cameras needed. The MG model, by its grid topology, is more prone to create fragmentation among the targets along the time, hence, leads to the creation of extra clusters for PFA and PIFA. LIFA is, however, not affected because it keeps the same clustering along the time. However, as shown in Figure 15(b), this impacts the capacity of coverage. The RPG, on the other hand, simulates a group behavior (i.e., each target belongs to a group and follows a logical center). The targets are prone to stay together and keep the initial number of clusters low and constant. Figures 17(a), 14(b) and 17(c) illustrate the time complexity with respect to the number of targets for RWP, MG and RPG mobility models. As shown, he LIFA has the lowest cost in terms of time complexity regardless of the mobility model. By its design, LIFA runs the fuzzy clustering (FC) algorithm only at the beginning and then updates only the way each cluster is covered by a camera. In Table 3, we summarize the results presented in Figs. 16 and 17 along with the complexity of the three algorithms.

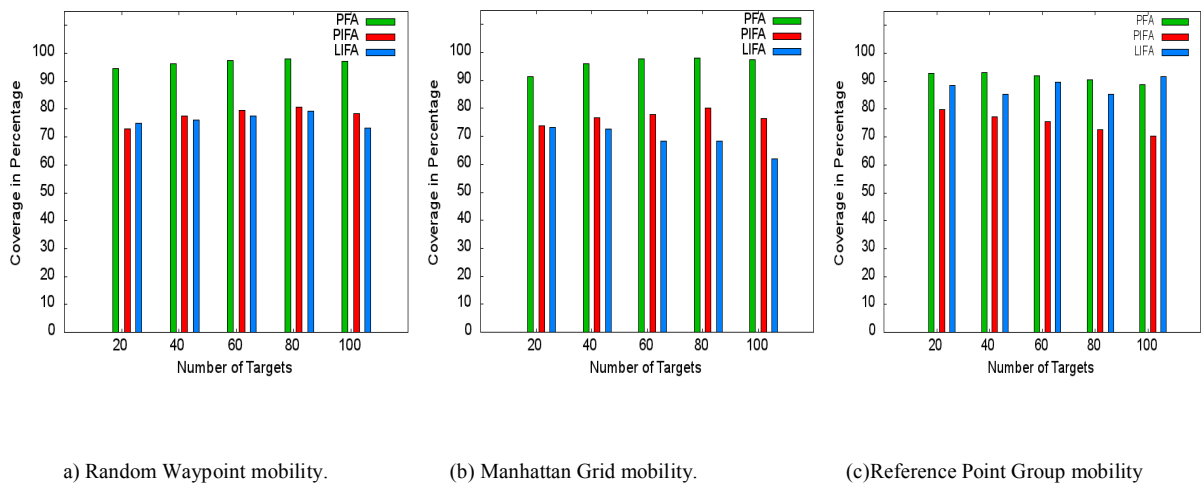

a) Random Waypoint mobility.　　(b) Manhattan Grid mobility.　　(c)Reference Point Group mobility

Fig. 15. Evolution of the average coverage ratio under different mobility patterns and with respect to the number of the targets (a, b and c).

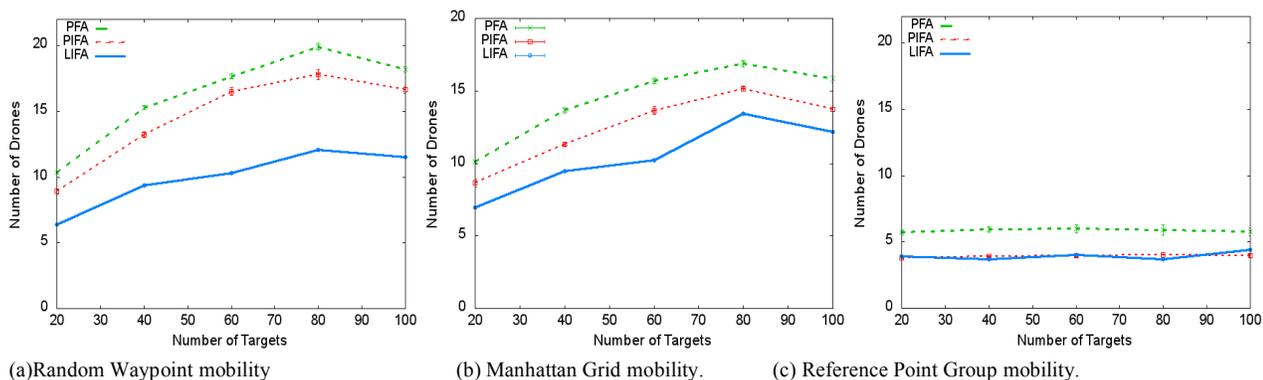

(a)Random Waypoint mobility　　(b) Manhattan Grid mobility.　　(c) Reference Point Group mobility.

Fig. 16. Evolution of the average number of mobile cameras required under different mobility patterns and with respect to the number of the targets (a, b and c

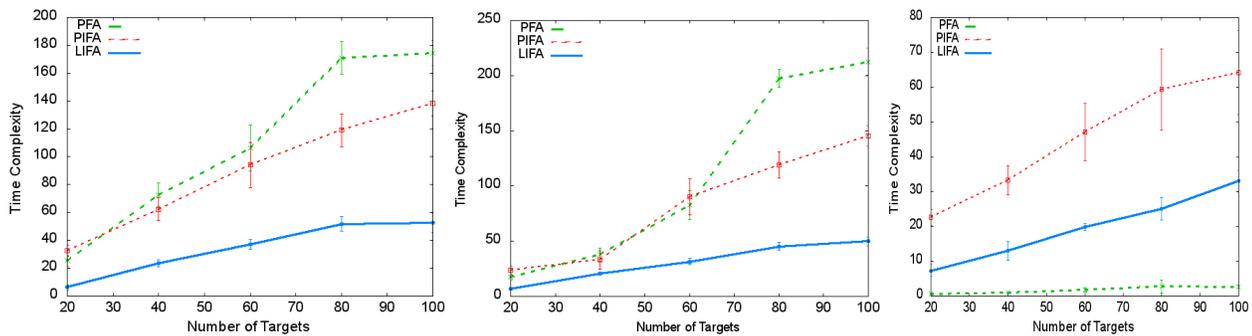

(a) Random Waypoint mobility.  (b) Manhattan Grid mobility.  (c) Reference Point Group mobility

Fig. 17. Evolution of the average time complexity with respect to the number of targets.

TABLE 3
COMPARISON BETWEEN PFA, PIFA AND LIFA

| | PFA | PIFA | LIFA |
|---|---|---|---|
| **Complexity** | | | |
| | $o(f(\hat{N}/\Delta t))$ (normalized over time) | $O(f(N)/\Delta t + N2\pi/\Delta\phi)$ (normalized over time) | $O(f(N)/t_{max} + N2\pi/\Delta\phi)$ (normalized over time) |
| **Number of drones** | | | |
| **Random way Point** | Highest | Middle | Lowest |
| **Manhattan Grid Mobility** | Highest | Middle | Lowest |
| **Reference Point Group Mobility** | High | Low | Low |
| **Time Complexity** | | | |
| **Random way Point** | Highest | Middle | Lowest |
| **Manhattan Grid Mobility** | Highest | Middle | Lowest |
| **Reference Point Group Mobility** | Lowest | Highest | Middle |

## C. Lessons learned

UAVs can be used for CPS applications to provide quick, low-cost, or non-lasting visual sensing solutions that are significant in real-world applications such as border protection and disaster recovery. Target and area coverage in such CPS applications are thus critical. To examine the design challenges in detail, we classify the problem of coverage in CPS surveillance applications into area coverage and target coverage. The former aims to cover an entire area regardless of the objects/targets inside, which is usually practical when monitoring is conducted in a controlled or reasonably small (dense) area. The work can be further classified into fixed camera deployment and area scanning. The latter can be expressed as reaching the minimal number of cameras and optimal camera placements and/or orientations using any directional sensor such that each target is fully or partially covered by at least one camera.

Although camera-based coverage issues in UAVs can be treated similarly to common coverage problems, certain particularities are related to UAVs. Initially, UAVs scan from the air, thereby creating a trade-off between having images/videos captured from a high altitude with a wide camera coverage area but producing low-quality images/videos, or having the image/video captured from a low altitude with a narrow camera coverage area but producing high-quality images/videos. Another trade-off is that UAVs are mechanical objects flown into the air; when the target is quite agile, having a UAV that is as quick as the target area be unmanageable. This situation resulted in the issue of designing, powerful, flexible, and lightweight UAVs while emphasizing the practicality of target mobility predictions. Another challenge is that sensors and cameras may be placed in UAVs at various positions [102]. Therefore, numerous structures and situations of the camera coverage area relative to the UAVs are likely. Their positioning during flight may change as well. Moreover, UAVs, particularly fixed-wing UAVs, may navigate the target area at high speed, consequently introducing coverage issues that include non-convex or disconnected areas [103]. Real localization and tracking investigations also showed that the UAVs and targets that move at high speed can cause unstable communication connections and packet losses. This issue may obstruct sufficient data collection, such as receiver signal strength (RSS) and time of arrival or data with several types of bias [104]. Table 4 maps the area and target coverage design challenges and requirements with CPS applications.

TABLE 4.
MAPPING UAV AREAS AND TARGET COVERAGE USE CASES, DESIGN CHALLENGES, AND REQUIREMENTS WITH CPS APPLICATIONS

| Applications | Use Cases, Design Challenges, and Requirements |
|---|---|
| **Transportation** | • Designing UAVs with effective camera-based coverage is significant for use cases such as monitoring traffic, tracking and reporting traffic violations, smart parking, and smart navigations.<br>• In transportation applications, mobility predictions can be highly crucial since cars work in deterministic directions.<br>• Speed of the cars may pose some challenges to the UAVs camera-based coverage.<br>• Intelligent transport systems may have several numbers of targets that should be effectivity covered by UAVs camera.<br>• The camera used in UAVs are commonly sensitive to occlusions; hence, in geometrically complex environments such as dense urban areas, this application becomes a more challenging task [105]. |
| **Constructions and Infrastructure inspections** | • Designing UAVs with effective camera-based coverage has several vital roles in use cases such as tracking real-time construction progress, producing 3D models for the planning and construction phases, inspecting existing structures, and ensuring the safety of the labors.<br>• UAV-based coverage is probably more suitable for large industry sites with large-scale infrastructure such as gas pipelines whose total line length is too long. Thus, effective camera coverage in such environments becomes more challenging [106].<br>• Robust collaborative scheme between various UAVs is highly required to provide effective camera-based coverage for such long-distance applications.<br>• The approach should be able to generate semantically rich 3D representations from captured overlapping images [28]. |
| **Surveillance** | • Designing UAVs with effective camera-based coverage is important to achieve several use cases such as crowd surveillance, border surveillance, and disaster surveillance.<br>• Using multi-UAVs for surveillance is becoming popular because these UAVs can reach places that were previously impossible to reach. However, in such environments, the required area of coverage may not be achievable.<br>• The approach needs effective data fusion schemes to provide enriched data about the area to be covered. |
| **Delivery of goods** | • Designing UAVs with effective camera-based coverage can have important role for goods delivery use cases such as providing safe UAVs landing, and ensuring successful doorsteps delivery.<br>• Area coverage can be crucial in the applications that use UAVs camera for constructing UAVs paths to deliver the goods.<br>• Delivery of goods involving payloads that may limit the use of cameras and other sensors onboard.<br>• The payload size may add another restriction to camera installment that may affect angle of views and the camera coverage area. |
| **Wireless & cellular systems** | • Effective camera-based coverage can be useful for use cases such as identifying areas of concentration to guide the deployment of base stations according to the expected traffic load.<br>• The size of the communication payload installed on UAVs to provide on-the-fly communication may restrict the installment of a camera if camera-based coverage is required. |
| **Medical and Healthcare Systems** | • Effective camera-based coverage UAVs can be significant for several use cases such as monitoring patients' movements using dimensional target tracking to avoid adverse events and emergency situations.<br>• UAVs can be used in medical facilities to track patients' movements. However, such application requires effective 3D target coverage.<br>• Using multi-UAVs for medical systems is important for reaching various locations that were previously hard to reach in a timely manner, such as remote and isolated locations and villages. However, in such complex and unattended environments, achieving practical camera coverage becomes more challenging. |

## IV. PATH/TRAJECTORY PLANNING

Having many UAVs require optimal path planning strategies. Examples of multiple UAVs performing several missions can be found in [107, 108] In order to save energy and lower the system latency, there should be a well-designed algorithm that prolongs the system life time and minimizes the system response as a result of failure events. We have to decide which UAV, how and when should go to a position. Upon which, there should be some swarm management and collision avoidance plans.

Planning the paths that UAVs go through to cover key areas and objects on-site is very crucial to minimize the aerial mechanical energy consumed for navigation, while insuring that the UAVs can fly in an obstacle-free plane, and avoid collision amongst each other. Authors in [109] and [110] propose path planning techniques based on re-optimization and graph-based techniques, respectively. However, they assume obstacle free paths, and do not address the possibility of collision amongst UAVs, which are two possible scenarios likely to happen while covering harsh environments.

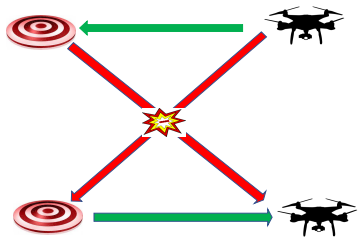

Fig. 18. The effect of proper path selection on collision avoidance.

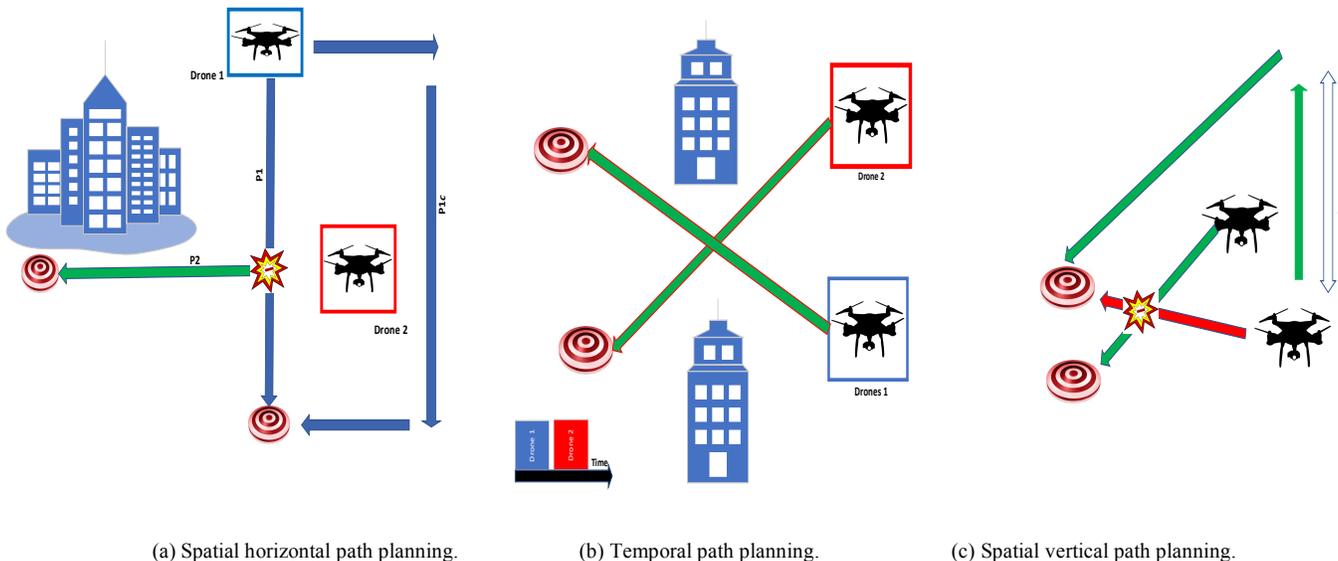

(a) Spatial horizontal path planning.　　　　(b) Temporal path planning.　　　　(c) Spatial vertical path planning.

Fig. 19. Degrees of freedom for collision avoidance.

### A. Collision avoidance

In a real-life surveillance problem, we are normally interested in monitoring a vast area encompassing many targets using multiple UAVs in general. Consequently, in an air space with several drones, there always exists the probability of collision, independent from the fact that the targets could be mobile entities. This is because sudden movements of the target could affect the whole system's behavior, or at least part of it momentarily. This is where collision avoidance issue becomes more challenging and vital. Generally, one could think of different ways to avoid two or more drones crashing into each other, either through spatial path planning in the same 2D plane, temporal path planning through controlling the time when drones can move, or through controlling the altitude of the drones in 3D (i.e., different altitudes). As a result, intelligent UAV selection to cover a target could sometimes eliminate the chance of such accidents (See Fig. 18). In general, considering delay and energy consumption, it seems logical to let the UAV, which is nearer in distance to a target to cover it. However, following this simple method may not be always fitting. Fig. 19(a) demonstrates how the collision is avoided by path changing. In this example, drone1 wants to get to its below target through path P 1, while drone2 wishes to cover the target on its left side through path P 2. While taking these straight lines could solve the problem readily and logically, there is a great chance of collision in the intersection of the two straight arrows. However, using spatial horizontal planning, if drone1 takes path P 1c in the same 2D plane as depicted, the collision is simply avoided. It is notable that in that case, the target is monitored with a slight delay compared to P 1, in addition to slight growth in power consumption since P 1c is longer.

Similarly, one could think of temporal planning as a technique for collision avoidance, through letting the drones to move in different time instances. Fig. 19(b) depicts time division-based target tracking where drone1 has the primary right of moving for the first-time instance. Note that, the drones may not cover the targets on their left due to having buildings occluding the line of sight. Alternatively, drones could fly a shift in altitude with respect to each other so as to lower the chance of accidents like in Fig. 19(c). Using the three degrees of freedom above, path planning problem could be mapped into a non-linear optimization problem that are usually solved by efficient distributed techniques subject to a specific goal. This could be the minimization of the total energy consumption by all collision free paths, providing the fastest performance to facilitate the real-time autonomous navigation of the UAVs for target tracking. If the optimization-based algorithm turns to have high complexity, sub-optimal algorithms like greedy algorithms (e.g., select shortest path first) could be considered.

Another challenge for UAV deployment is collision with moving or stationary objects. In this case, the collision avoidance system comprises a sensing and detection step and a collision avoidance and maneuver step. In the first step, Automatic Dependent Surveillance-Broadcast (ADS-B) system, in which the UAV periodically transmits information about its altitude, speed and location through GPS to a ground control station [111], [112]. Other sensing methods include radar [113], [114] and infrared sensors [115]. Collision estimation techniques include distance-based, act-as-seen [116], probabilistic estimation [117] and worst case estimation [118]. In the second step, once a potential collision is detected, a collision avoidance technique is applied in order to maneuver around the possible collision position. The most popular collision avoidance techniques include geometric technique [112], [119] which exploits location, speed and direction of the UAVs to estimate collision trajectory in order to maneuver away from it. Optimized trajectory technique is proposed in [120], [121], which is similar to the geometric technique, but there, the collected information is processed to generate an optimized trajectory through solving an optimization problem, which leads to higher computational complexity. The bearing angle-based technique is proposed in [122], [123], in which a visual sensor is used to calculate the relative angle between the obstacle and the moving UAV in order to ensure that the obstacle's image remains in a safe position from the camera's FoV. For further details on collision avoidance techniques, we refer the readers to [118], [124-129] and the references therein.

### B. Swarm Formation

Formation control of multi-robot systems performing a coordinated task has become a challenging research field, mainly for military applications. The formation problem is described as obtaining a control algorithm that guarantees that several autonomous vehicles can uphold a particular formation while traversing a generalized trajectory (i.e., not necessarily linear) and avoiding collisions simultaneously [130]. Such an algorithm, however, may require global picture of all UAVs, which may affect the scalability. Also, it is recommended for an effective swarm to be able to recover and reconfigure itself in case of failure, or incremental events such as member add/remove. In this way, the swarm could be a reasonable answer to the fast-moving targets in terms of coverage.

Different algorithms have been developed to control the formation and reformation of a swarm of UAVs, which somewhat correspond to the team orienteering problem (TOP) [131]. In[131], a genetic algorithm and ant colony optimization hybrid algorithm (GA-ACO) is proposed to solve the multi-UAV mission planning. In [132], optimal resource allocation using Evolutionary Game Particle Swarm Optimization (EGPSO) with the *2-D* geometric path planning algorithm based on Pythagorean Hodographs (PHs) is proposed. In [109], the authors construct a feasible path using Genetic algorithms and then smooth this path using Bezier curves. In [87], the authors consider dynamic path planning, where targets could be added or removed at different time slices, thus transforming the dynamic UAV path planning problem into a series of static problems. The dynamic branch and price algorithm with a periodic re-optimization framework are used to solve the multi-UAV path planning problem. In [133], each UAV is assigned some control points to cover and the path is then planned using Voronoi graph and Dijkstras algorithm. In [134], a fuzzy virtual force approach is proposed to the problems of collision avoidance and simultaneous arrival in multi-UAV path planning, where some UAVs might actually have to take a longer path than intended just to ensure simultaneous arrival.

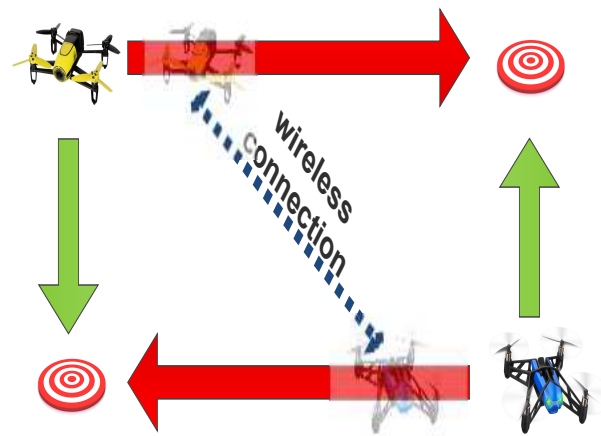

Fig.20. Proper path selection to save energy

### C. Energy Planning and UAV energy profiling

One important ingredient for path planning is UAV energy profiling. Energy profiling focuses on designing accurate models that can be used to estimate the aggregate energy consumption as a result of certain actions such as flying up, down, or horizontal for a particular distance as part of target detection and/or tracking. Also, for some applications, it may be required to estimate the energy required to capture/process data using camera, and other orientation sensors integrated as part of the UAV platform. According to the experimental results, almost about 90%-95% of the drone' energy is consumed by its mechanical movements. Therefore, there should be proper energy models that could be used to estimate the energy of the drone for efficient coverage purposes, because each extra movement of the drone costs energy. So, one sight of the design could be to lower the extra movements of the drones on the way to their corresponding targets. For example, in Fig. 20, one could suggest that the drones take the vertical (green) paths instead of the horizontal (red) ones to save energy. However, considering other factors such as wireless network connectivity, taking the horizontal paths might be better in terms of wireless communication between the participating drones to enhance the coordination and facilitate the target coverage. Such a trade-off requires accurate UAV energy models and profiles, in addition to wireless channel models for drone navigation in order to quantify the energy consumption as a function of different system parameters, integrating coverage accuracy with efficient communication. In [135], a theoretical model is derived for the energy consumption of a fixed-wing UAVs as a function of the flying velocity and acceleration. Taking into consideration the flight radius and speed, the authors proposed a circular UAV path, that maximizes energy efficiency. They also developed an algorithm that finds the optimal path subject to general path constraints including acceleration, range of speed, initial, and final locations. The authors in [136] present a study on the tradeoff between power consumption and flight performance, taking into consideration the UAV's stability while comparing different controlling techniques.

On the other hand, the multi-UAV path planning is a process where UAVs try to find a collision-free path from their starting points to their destinations simultaneously, while minimizing the total mechanical energy. To achieve this, we conducted a comprehensive set of experiments to estimate the energy consumed by the UAV based on moving horizontally, or vertically upwards and downwards. This is in addition to performing communication functions, such as video streaming, sending sensor measurements, etc. The UAVs battery have been connected with a DC power supply as shown in Fig. 21. The voltage of the power supply is kept fixed to that of the drone's OEM (Original Equipment Manufacturer) battery specifications, 30 V. Afterwards, the current I required for some of the UAV movements is measured, and multiplied by the voltage to get the electrical power in watts as shown in Table 5. III. It is worth mentioning here that the mechanical energy consumed for aerial control and navigation is the predominant energy by far compared to the processing or communication energy. Hence, it is the most critical to be minimized in order to maximize the UAV flight time.

Most of wireless nodes utilize a stand-alone battery unit that in, frequent replacement of the batteries is the main irritating problem. This problem seems more crucial when the batteries are installed in a dangerous environment like toxic areas. Therefore, many efforts to prolong the lifetime of communicating nodes have been made [137] and several lifetime maximization techniques have been proposed [138]. In

[139] node cooperation is applied to optimally schedule the routing in order to minimize the energy consumption and the delay. The power efficiency problem is another major concern for wireless video surveillance (WVS) applications. A large body of research has focused on energy conservation in WVSNs. Each, help the node's battery to stay alive for a longer period of time. Nevertheless, the issue with battery replacement still remains unsolved even with the help of the mentioned mechanisms.

As an attractive and practical alternate solution to the above highlighted issues, Energy Harvesting (EH) nodes have gained considerable attentions due to their potential unlimited energy supply capabilities by physical phenomena [140-142]. Energy is harvested by solar, vibration and thermal effects or any other physical phenomena that could act as an energy source. A review of the EH techniques is found in [140]. A promising harvesting technology is the RF energy transfer where ambient RF radiation is captured by the receiver antennas. By using appropriate circuitries [143],[144] the radiations are converted into a direct current voltage that can power up a wireless node. Wireless power transfer (WPT) or wireless energy transmission is the transmission of electrical power from a power source to a consuming device without using discrete manmade conductors. Wireless transmission is useful to power electrical devices in cases where interconnecting wires are inconvenient, hazardous, or are not possible. In wireless power transfer, a transmitter device connected to a power source transmits electromagnetic waves within an intervening area to receivers, where this electromagnetic waves are converted back to electric power and utilized [145]. Wireless power techniques can be categorized into two main categories namely non-radiative and radiative. In non-radiative techniques also referred to as near-field techniques power is transferred over short distances by magnetic fields using inductive coupling between coils of wire or in a few devices by electric fields using capacitive coupling between electrodes [146]. Of course, this category has got nothing special for UAV systems due to the fact that UVAs are mobile and basically could go so far away from the RF generator. In radiative or far-field techniques, also referred to as power beaming, power is transmitted by beams of electromagnetic radiation, like microwaves or laser beams. These techniques can transport energy longer distances but must be aimed at the receiver. The observation field could be equipped with a WPT antenna to charge up the UAVs battery and thus prolong the lifetime of them. Clearly, we could not neglect the fact that the amount of absorbed energy could not be so much to supply all the need of a drone in terms of mechanical, circuitry and communications specially, when the distance grows high. So, it is expected that it could compensate some portions of the energy consumed by the drones. Practically, the absorbed energy will not be comparable to the mechanical losses therefore, the method of RF energy harvesting could be more logical for fixed point sensor nodes. It is worth mentioning that, instead of using low efficient RF energy harvesting techniques for moving UAVs, one could think of deploying solar panels installed on the top of the UAVs. Although, the sun shine is not available during the whole day or basically in some areas based on their geometry, they could be implemented while the size, weight and the stability of them are carefully chosen.

Simultaneous wireless information and power transfer (SWIPT) on the other hand, could be another solution to the energy shortage in the UAV systems that include fixed sensor nodes. The idea suggests to utilize a single antenna for both information and energy transfer. A time instance is divided into two parts. A portion for information transfer and the rest for energy transfer (See Fig. 22).

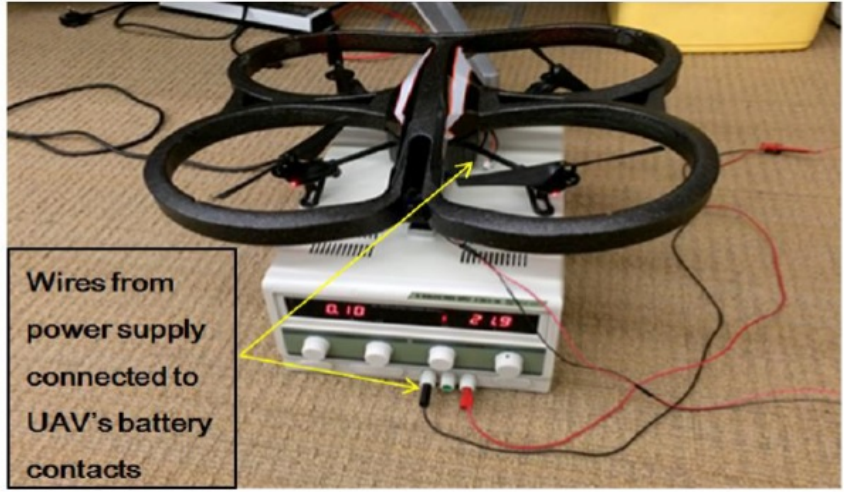

Figure. 21. AR Drone connected to power source to measure the energy required for some UAV movements.

TABLE 5

POWER ENERGY CONSUMPTION OF DIFFERENT DRONE MOVEMENTS

| Movement | Power Consumption | Energy Per Meter |
|---|---|---|
| Horizontal | 12W | 2.4J |
| Vertical Upwards | 15W | 3J |
| Vertical Downwards | 13.5W | 2.7J |

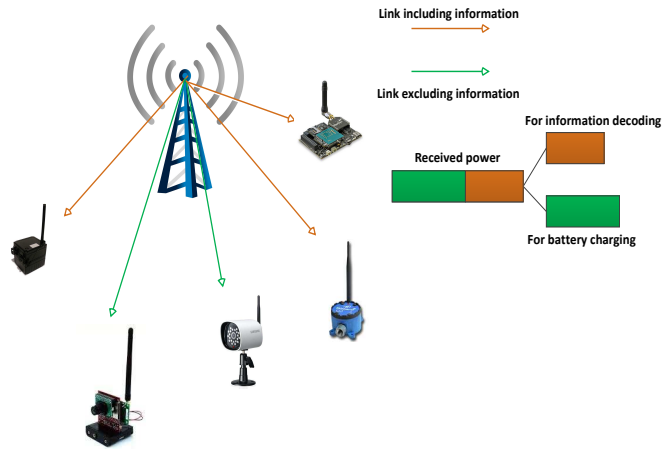

Fig. 22. Using simultaneous wireless information and power transfer in UAV system

### D. Lessons learned

#### 1) Collision avoidance

The effective integration of UAVs in several CPS applications must ensure that UAVs are capable of flying safely within commonly shared flying systems, spaces, and environments. A vital factor in achieving such integration is collision avoidance. The functionality of collision avoidance in UAVs systems is to guarantee that no collisions occur with other UAVs and/or the targets [127]. To realize an operational collision avoidance scheme, the following design challenges should be considered [127]. First, appropriate methods should be provided that can sense the environment and obtain essential data, such as size, speed, geo-location of other UAVs, targets, and obstacles. Second, efficient methods that can utilize the collected information in a timely manner must be provided for deciding when and how to apply or stop collision avoidance operations. Third, appropriate design requirements must be determined, such as type of UAV, power consumption, and payload, as well as whether sensing dimensions are 2D or 3D and whether the target is a moving or stationary object.

#### 2) Swarm formation

In CPS applications, multi-UAVs are more desirable than a single UAV system. Particularly, multi-UAVs operate jointly to achieve tasks. Concurrently, additional missions can be delivered to decrease the risk of failure due to mobile UAVs or component failures [147]. Multi-UAVs are essentially required to operate in cooperative relationships [148]. The formation flight control of multi-UAVs is a significant direction. The key goal is to organize a swarm of UAVs to accomplish a required formation, while avoid intervehicle collisions. [147-149]. Given that formation flight is a vital cooperative way in several circumstances, the idea of formation path planning is to solve the multi-UAV path planning problem. In particular, formation path planning is proposed for each UAV to be able to discover its collision-free path and maintain its formation organization. Therefore, to solve the problem in multi-UAV formation path planning, the following needs to be considered: path planning of a single UAV, construction of a suitable formation organization, and regulation of the reformation [147, 148].

In reality, CPS applications have high-mobility environments that can create highly dynamic network topologies of UAVs that are generally lightly and alternatingly connected. Therefore, efficient multi-UAV organizations and UAV swarm operations should be proposed to ensure reliable network connectivity [33]. Moreover, several previous studies have proposed solutions based on the assumption that multi-UAVs fly in the same plane. However, the problems of complex assumptions, such as 3D formation control for UAVs, need further investigation in the future.

#### 3) Energy planning

The requirement of several CPS applications can vary from one application to another; consequently, the mission of UAVs can also differ. UAVs are energy-constrained devices with a limited quantity of on-board energy for communications, mobility, control, processing, and payload. The factors that control power consumptions vary from one application of CPS to another, thereby making energy planning a challenging task. The energy constraints of UAVs must be considered because they notably influence the realization of effective UAV systems. The limited on-board energy of UAVs is also a limitation for the implementation of UAVs in various CPS applications[17].

The roles of multi-UAVs require an effective energy management that is reliable and utilize smart energy systems that can assist UAVs in accomplishing their functions successfully. Future directions should involve further investigations on employing wireless charging schemes because progresses in this technology are estimated to have a potential impact for UAVs. Another future direction is to utilize computational intelligence methods, such as deep reinforcement learning, in planning the path and battery scheduling for reducing the required energy to accomplish missions [1]. Table 6 maps the challenges and requirements of UAV path/trajectory planning design with CPS applications.

TABLE 6

MAPPING UAV PATH/TRAJECTORY PLANNING USE CASES, DESIGN CHALLENGES, AND REQUIREMENT WITH CPS APPLICATIONS

| Applications | Use Cases, Design Challenges and Requirement |
|---|---|
| Transportation | • Designing UAVs with effective path planning is significant for several use cases, such as delivering real-time monitoring of the vehicles in dense urban and high-mobility environments, and flawlessly collisions avoidance in congested locations.<br>• Intelligent transport systems involve dynamic environments with high-speed moving targets, deeming UAV-based monitoring such a challenging task.<br>• Developing precise coordination algorithms for UAVs system is a challenge that should be considered to enable UAVs for intelligent transport systems [105].<br>• Dense urban locations need real-time responses to avoid collisions; consequently, minimizing delays in planning UAV paths should be considered as one of the essential requirements.<br>• An effective energy management is required to maximize the UAV flight time for sufficient data collection. |
| | • Designing UAVs with effective path planning is important for use cases such as long-distance infrastructure inspections, and Multi-UAVs for monitoring construction sites [106]. |

| Constructions and Infrastructure inspections | • In large surfaces, constructions and infrastructure inspection using multi-UAVs can be an effective solution. However, an effective multi-UAV coordination method is required to maximize the efficiency of the multi-UAV members. An appropriate synchronization should be verified among all UAVs to accomplish their mission successfully.<br>• Real-time responses are required to avoid collisions; consequently, delays must be at the minimum. In such case, collision avoidance is addressed with regards to collision with existing infrastructure as well as inter-UAVs collision.<br>• An effective energy management is required, particularly for long-distance missions.<br>• Single and multi-UAVs may move with different speeds; therefore, effective path planning methods that consider the diverse motion requirement are required |
|---|---|
| Surveillance | • Designing UAVs with effective path planning is significant for several use cases such as large area monitoring, reaching places that were previously hard to reach.<br>• The area to be scanned in surveillance applications can have an arbitrary number of targets.<br>• Using multi-UAVs in surveillance applications has a high accidental rate [150].<br>• Although UAVs can reach places that were previously impossible to reach, such place increase the number of uncertainties.<br>• Real-time responses are required to avoid collisions.<br>• Surveillance of several UAVs should be performed to achieve the required mission; therefore, precise collaboration and coordination algorithms for multi-UAVs are necessary.<br>• Effective energy management is required, particularly for large-area scanning missions. |
| Delivery of goods | • Designing UAVs with effective path planning is significant to provide collision-free path for goods delivery, and to ensure a safe delivery of goods.<br>• Delivery of goods involves important factors that need to be considered while designing navigation methods, namely, dynamic weight and payload size.<br>• Delivery of goods requires a safe departing and landing scheme, particularly if the delivered goods are fragile.<br>• Delivery of goods in dense urban locations needs real-time responses to avoid collisions.<br>• Effective energy management is required, particularly if the delivery address required flying long distances.<br>• Precise coordination algorithms are required, particularly with the expected rapid growth of several UAVs for delivering goods [105]. |
| Wireless & cellular systems | • Designing effective path planning for UAVs has significant role for use cases such as ensuring the effective deployment of UAV-mounted base station.<br>• Designing an optimal path planning for wireless network applications should consider improving the key QoS metrics of communication systems, such as coverage probability, throughput, and delay.<br>• Given the number of users covered that have a UAV as base station, an accurate planning for the UAV in base station operations is further required [151].<br>• UAV placements and movements should consider UAV energy constraints.<br>• The trajectory of UAVs is considerably subjected to other factors such as flight time, ground users' demands [152], quality of experience [153], and collision avoidance [17]. |
| Medical and Healthcare Systems | • Designing effective path planning for UAVs is crucial for use cases such as ensuring real-time response in emergency cases, and ensuring the safety of chemical materials delivery.<br>• Medical and health applications can involve delivering sensitive material; therefore, such factor should be considered in path planning design.<br>• Real-time response is important because this application is related to human life.<br>• Effective energy management is required, particularly for long-distance missions. |

## V. IMAGE ANALYSIS AND VISION-BASED TECHNIQUES

As an intuitive representation of the objects under investigation, images are the most popular choice for UAV applications. Image analysis and vision-based techniques are typically needed for target detection, localization, and possibly target tracking. Advances in signal processing, image/video processing, statistical analysis, machine learning, and computer vision are needed to accurately and automatically process UAV-sourced imagery data, which is key to achieving the promise of UAVs. Some common problems in UAV-sourced data processing include photography, image matching, mosaicking and reconstruction, and decision making (e.g., classification and recognition).

CPSs are highly dynamic environments with a large number of components that require flexible, autonomous multi-UAV systems to provide successful monitoring, and tracking approaches. Such environments can rise several issues related to image analysis in UAV systems, such as 1) dynamic background due to UAV movements, 2) image de-noising to remove artifacts due to UAV mobility, 3) image stitching techniques from multiple UAVs.

The following subsections discuss Multi-UAVs for visual monitoring of CPSs and its related lessons learned.

*A. Multi-UAVs for visual monitoring of CPSs*

Regarding the CPS context, the use of multi-UAV systems for visual monitoring can hold significant practical applications. The analysis of a large number of images and video can be potentially used for regular inspecting manufacturing sites, monitoring large areas such as oil/gas pipelines, supervising work-in-progress, and checking existing construction. The importance of using multi-UAV systems for visual monitoring increases specifically in areas that are difficult to reach or large to monitor.

The significant processes for successful visual monitoring require effective executions of the following processes [28]. First, equipment and strategies capable of collecting informative visual data must be acquired; second, robust visual data analysis algorithms must be presented; and third, suitable information visualization methods must be obtained.

*1) Collection of informative visual data*

Collecting sufficient and high-quality data is a key factor to construct an accurate visual monitoring and detecting model. Multi-UAVs are required to gather representative data in the form of images and videos for a specified CPS. To have effective collection methods, the following should be considered:

- Identifying the requirement of CPS application and selecting the appropriate collecting components that balance the required quality and quantity of the collected data and implementation costs.
- Providing data collection strategies for multi-UAVs that allow effective collaboration among UAVs during data collection.
- Identifying the most informative views for data collection.

*2) Effective visual data analysis algorithms*

Visual data analysis has been a research subject for many years, and several image and video processing techniques have been proposed. However, with the advancement of computational power in recent years, deep learning networks have accomplished accuracies that are far beyond those of classical methods in object detection and image analysis [154, 155]. Using learning algorithms for UAV applications in CPSs to process real-world data intelligently can potentially extend the uses of UAVs in CPSs for a wide range of applications. Learning algorithms likewise have potential uses toward effectively utilizing the advantages and advancing the uses of multi-UAV systems for CPS applications. The potential uses of learning algorithms in multi-UAV systems for CPS applications can be summarized as learning algorithms for target detection and tracking, navigation, and planning.

One of the most effective learning-based paradigms is deep learning algorithms. Deep learning networks are one of the most powerful that yield state-of-the-art performances in many recognition and classification tasks, such as image recognition and segmentation, object detection and localization, speech recognition, and text recognition. Deep learning is inspired by the human brain's deep architecture, and deep learning models usually consist of deep stacked layers of neurons. In image-related tasks, a particular type of deep learning model, namely, deep convolutional neural network (DCNN), is shown to achieve state-of-the-art performances in many image problems [156-159]. DCNNs introduce a special type of neuron, namely, convolutional layers, to better discover the spatial structure of images. With image raw pixels as inputs and given target outputs in the training data set, DCNNs iteratively update the weights in all layers in a supervised training fashion. For large-scale tasks (e.g., ImageNet challenge with over 1 million data records [157]), Stochastic gradient descent is used to update weights after scanning a small number of data (mini-batch) to train DCNN models efficiently [160].

Deep learning methods applied on images captured from UAVs have an actual significance in CPSs. For instance, previous studies proposed the deep learning approach for detecting and counting cars for images captured from UAVs [161], road traffic monitoring from UAVs [162], and inspecting civil infrastructure systems [28, 163].

Real-time object detection should be considered to utilize the full benefits of the advancement in deep learning for image analysis. The reason is that this detection has a potential and wide range of applications for CPSs, such as infrastructure inspection, surveillance, and search and rescue. To meet this requirement, several studies have recently developed cloud-based deep learning to support the image analysis of UAV systems. For instance, CNNs have been used to enable UAVs for detecting several numbers of object categories. To eliminate the computational restriction of CNN, a hybrid method was previously proposed: this method moves recognition into a computing cloud but keeps low-level object detection and short-term navigation onboard [164].

*3) Information visualization and augmented reality methods*

Considering the application of multi-UAVs in CPS, information visualization methods play an important role in utilizing UAV features. The visualization methods are also essential for visualizing and discovering various modalities of user interaction with real data from physical application, thereby providing numerous advantages. A promising visualization technology that can be integrated to multi-UAV systems for CPS applications is augmented reality (AR). AR is an interface that links digital information to the user's view while spatially aligning to the current physical environment. This type of information presentation is relevant for various professional tasks. AR has been successful for several industrial applications [165].

Integrating AR to multi-UAVs can further enhance their application in CPS. For instance, AR can provide real objects as the UAVs capture images but also additional important and informative images, text, or marks over these images. Hence, the image captured by UAVs becomes well informative. However, only few studies have investigated the integration of UAVs and AR technology. For instance, the AR display and control system of a data-driven UAV ground station have been designed. Another study [166] showed that the videos of UAVs with a video camera can be adjusted with the 3D viewport for AR display[167]. The potential of AR to transform the UAV industry is rapidly being recognized, and its application for CPS as a result is enhanced and improved. The collaborative combination of AR and UAVs can produce exclusively novel industry applications. AR-enabled UAVs hold practical significance and motivate further researches. Table.7 presents application objective, challenges, recent advances, and future directions on vision-based analysis in UAV systems for CPS applications. summary and comparisons of recent studies that use vision-based analysis methods in UAV systems for CPS application. Table.8 provides summary and comparisons of recent studies that use vision-based analysis methods in UAV systems for CPS application.

TABLE 7

APPLICATION OBJECTIVE, CHALLENGES, RECENT ADVANCES, AND FUTURE DIRECTIONS ON VISION-BASED ANALYSIS IN UAV SYSTEMS FOR CPS APPLICATIONS

| vision-based analysis methods | Application objective | Challenges | Recent Advances and future directions |
|---|---|---|---|
| **Object Detection** | • Target recognition<br>• Target tracking | • Long distances, small targets, incomplete view, and diverse and dynamic environments pose difficulties.<br>• Images captured by UAVs are deteriorating from illumination, rotation, and scale changes.<br>• Meeting real-time tracking requirement is difficult, particularly when tracking a target moving at high speed.<br>• Object detection methods mostly are proposed to specific objects and contexts<br>• Object detection is an expensive computational task. | • Proposing image analysis methods based on cloud computing and developing distributed algorithms to support real-time processing<br>• Proposing robust analytical methods, such as deep learning, which can provide high-accuracy object detection.<br>• Capturing and handling multi-view image<br>• Proposing cooperative scheme for multi-UAVs system that utilize effective image stitching techniques.<br>• Proposing vision-based methods that can handle incomplete views<br>• Using multiple sensors and data infusion along with image analysis methods |
| **Localization and Mapping** | • Path planning<br>• Context awareness<br>• Flight control | • In practice, computational cost with dense mapping is expensive.<br>• Localization and mapping in UAVs moving at high speed can be a challenging task.<br>• Localization and mapping in dynamic and unstructured environments require sophisticated methods. | • Employing sophisticated vision-based methods for localization and mapping, and proposing strategies for deployment within power-constrained devices such as UAVs<br>• Ensuring real time response when vision-based methods are used for localization and mapping. |
| **Monitoring** | • Site monitoring<br>• Safety inspection | • A real-time analysis and response are required.<br>• Computational cost is expensive.<br>• Monitoring based on visual data analysis in dynamic unstructured environments may suffer from several unpredictable situations. However, since UAVs are flying vehicles, UAVs are less tolerant with dynamic uncertain environment. | • Using image analysis methods in several monitoring applications in complex environments given the development of visual data analysis methods such as deep learning<br>• Using multi-view images, multiple sensors, and image fusion that can provide effective monitoring applications<br>• Integrating AR technology into multi-UAV systems that can provide informative and interactive interface for good monitoring approaches<br>• Provide an effective energy and battery charging method, particularly for applications that involve long distances |

TABLE 8

SUMMARY AND COMPARISONS OF RECENT STUDIES THAT USE VISION-BASED ANALYSIS METHODS IN UAV SYSTEMS FOR CPS APPLICATION

| References | Research area | Application objective | Potential Use in CPS | Type of Data | Vision analysis methods |
|---|---|---|---|---|---|
| [168] | Localization and Mapping | Navigation | Outdoor navigation | Image | Deep learning (CNN method) |
| [169] | Localization and Mapping | Navigation | Indoor navigation | Image | Deep learning (CNN method) |

| [164] | Object Detection | Multi-objective type recognition | General | Image | Deep learning faster regions with CNNs [170] |
| [171] | Localization and Mapping | Autonomous landing | General | Image | Unsupervised perception model |
| [104] | Object Detection | Recognition and tracking of multiple moving targets | Target tracking | Video image | Machine-learning-based model |
| [172] | Localization and Mapping | Estimation of translational position and velocity of UAVs | General | Image | Visual simultaneous localization and mapping algorithm |
| [173] | Localization and Mapping | Path planning | Search-and-rescue missions | Image | Deep learning (CNN method) |
| [174] | Monitoring | Construction monitoring | Civil infrastructure | Image with AR | Integrating UAVs with AR technology |
| [175] | Monitoring | Surveying sites | Civil infrastructure and industrial sites | image | Generation of 3D model from UAV images |
| [176] | Monitoring | Safety inspections | Civil infrastructure and industrial sites | Image | Statistical analysis |
| [177] | Object Detection | Object recognition using data-related vehicles and solar panel detection issues | General | Image | Convolutional support vector machine |
| [178] | Object Detection | Objective classification | Oil palm tree detection (Agriculture) | Image | Deep learning CNNs |
| [179] | Object Detection | Objective recognition | Agriculture | Image | Neural network architecture that uses histograms and convolutional units |
| [180] | Localization and Mapping | Path planning | Search and rescue (air–ground collaborative teams) | Image | Deep learning (CNN method) |
| [181] | Monitoring | Construction monitoring | Civil Infrastructure and industrial sites | Image | Generation of 3D model from UAV images |
| [182] | Localization and Mapping | Navigation | Indoor navigation | Image | Recurrent neural networks |
| [183] | Object Detection | Objective classification | Weed detection (Agriculture and security) | Image | Sparse autoencoder |
| [184] | Object Detection | Objective recognition and classification | Automated detection and classification of micro-UAVs | Image | Deep belief network |

## B. *Lessons learned*

As an intuitive representation of the objects under investigation, images are the most popular choice for UAV applications. Image analyses and vision-based techniques are typically needed for target detection, localization, and possibly target tracking. Ranging instruments, such as visible band (e.g., common and representative RGB cameras) and near-infrared cameras, are identified in [185] for UAVs. Advances in signal processing, image/video processing, statistical analysis, machine learning, and computer vision are needed to process UAV-sourced imagery data accurately and automatically, which is key to achieving the promise of UAVs. Table.9 maps the challenges and requirements of UAV vision-based techniques design with CPS applications.

TABLE 9

MAPPING OF USE CASES, DESIGN CHALLENGES, AND REQUIREMENT WITH CPS APPLICATIONS ON UAV IMAGE ANALYSES AND VISION-BASED TECHNIQUES.

| Applications | Use Cases, Design Challenges and Requirements |
|---|---|
| Transportation | • Designing UAVs system with effective vision-based techniques is significant for use cases, such as vehicles tracking, plate number recognition, and smart navigation.<br>• Require real-time detection, and tracking of cars for drivers' behavior analysis and violation recognition<br>• Require detection methods that can identify details of small images with high accuracy, such as the car plate number from images taken from high altitude with low-quality resolution<br>• Require methods that can handle images taken from UAVs moving at high speed, as well as with incomplete views, because intelligent transport involves high-speed moving vehicles<br>• Require high computational processing unit due to multi-object vision detection methods applied to urban areas that have several hundreds of objects |
| Constructions and Infrastructure inspections | • Designing UAVs system with effective vision-based techniques is important for several use cases such as automatic construction monitoring, 3D and 2D models reconstruction, infrastructure site evaluation, physical obstacle recognition, construction process tracking, and safety inspection.<br>• Requires vision-based methods to handle images that deteriorate from illumination variation and image viewpoint variation<br>• Requires real-time detection of failures or accidents in oil/gas pipelines lines, to avoid them at initial stages.<br>• Require effective methods that can deal with images taken form high altitude with low quality resolution<br>• Require high computational processing unit of multi-object vision detection methods applied for constructions and infrastructure inspections |
| Surveillance | • Designing UAVs system with effective vision-based techniques acts as operational core for use cases to achieve efficient object detection, real-time target tracking, and automatic monitoring.<br>• Human and context recognition and vision-based tracking for national security and intrusion detection purposes are amongst the top use cases in this category of applications. Consequently, this application requires to develop effective situation-aware surveillance methods<br>• Require finding alternative methods based on vision-based analysis motion control that can handle unknown and complex environments<br>• Require effective visualization methods to be integrated to surveillance systems<br>• Require effective methods that can deal with images taken from high altitude with low-quality resolution<br>• Require high computational processing unit |
| Delivery of goods | • Effective vision-based techniques are crucial for delivery of goods use cases such as real-time navigating and routing, and safe good deliveries.<br>• One of the vital challenges is that using UAVs for goods delivery may have limited use of computational resources and vision sensor on-board due to the payload of goods<br>• Require real-time response and high computational cost, though vision-based methods can be used for navigation systems for indoor application of deliveries of goods |
| Wireless & cellular systems | • Effective vision-based techniques can be useful for uses cases such as using UAVs to complement the cellular coverage through detecting and tracking crowds for effective deployment of base stations in near real-time.<br>• Involve several variables due to continuous UAV trajectories to be calculated, although the vision-based method can be used for path optimization [33]<br>• Vision-based techniques can be used in tracking crowds for effective deployment of base stations. However, such application requires effective vision-based methods that may require high computational power.<br>• May affect deployment of highly processing unit on-board due to the size and weight of the payload |
| Medical and Healthcare Systems | • Designing UAVs system with effective vision-based techniques is vital for use cases such as human tracking, identify mosquito habitats for eradicating malaria, and identify the critical case (e.g., drowning victims)<br>• Human tracking and behavior/context detection are crucial for identifying critical and adverse events. However such applications requires effective real-time analytical methods that can identify the critical case (e.g., drowning victims [186]) immediately<br>• Require effective methods that can deal with images taken from high altitude with low-quality resolution, such as using UAVs to identify mosquito habitats for eradicating malaria [37] |

## VI. NETWORKING AND CROSS-LAYER DESIGN FOR SCALABLE AND SECURE COMMUNICATION

Above all, networking issues are another important aspect of multi-UAV system. This in turn, could be divided into two main categories, namely; Wireless topology and connectivity control, and cross-layer design and quality of service of the MANET of drones.

### A. Network Connectivity and Topology Control

UAVs should stay connected to the system during their trips over an area. Otherwise, there is a great chance of losing connection to the fusion node and thus, not being able to give service any more. The only feasible case that the drones could lose the connectivity temporarily is that, when the drone is ordered to fly over an area without coverage and collect data then, come back to an area which is within the wireless network coverage and pass the collected data.

In the literature, the coverage and the connectivity optimization have been heavily investigated for sensor networks and mobile robots. The main focus of the studies conducted by the sensor network community is to save energy by scheduling activity (i.e., activate or de-activate a set of fixed sensors) while maintaining connectivity and area coverage simultaneously [73-76]. These studies only consider fixed deployment and anisotropic (directional) sensors. On the other hand, the connectivity is eluded and it is generally assumed that the number of sensors is known. The covering problem becomes how to move and direct sensors to optimize a coverage objective, such as an accumulation function [69]. In both cases, the coverage is understood as a flat coverage, i.e., the target is considered as covered if at least one sensor has the target in its sensing range. These typical criteria is not sufficient and some applications require a target to be covered from different angles [56].

To ensure coordination and intra-swarm navigation among deployed UAVs, several approaches of swarm models are presented in the literature [160, 187-190]. However, these approaches are meant to maintain certain formations among the UAVs with collision avoidance, but do not consider network connectivity nor distributed coverage. Hence, in certain environments such as metallic structures, urban environment or in areas with no or poor wireless connectivity, UAVs operation could be constrained by communication limits.

### B. Cross-layer design and Quality of Service (QoS)

Quality of service (QoS) plays an important role in accuracy and reliability of a sensor network from a different point of view. Like any other communication systems, here also the specific application dictates the necessary requirements that the system should reach. There may be some applications that demand higher bandwidth for being observed properly, or certain quality of video captured from the mobile cameras to facilitate efficient monitoring. Hence, adaptive multi-resolution imaging or/and compression techniques, leveraging mobile edge computing (MEC) paradigm [48], should be used so as to summarize the sensing and imagery data into the available bandwidth, resulting into minimum end-to-end delays. Of course, the amount of bandwidth taken up with the control information that are used for UAV's navigation is far less than the UAV's measurements and imaging.

Based on the QoS requirements, including throughput and delay for end-to-end data transport, the UAVs are required to schedule the data transport to the central server, possibly using multi-hop transmission through other UAVs or fixed access points. However, due to frequent mobility, and based on the current topology generated through swarm formation, the wireless channels quality intermittently changes, causing interrupted communication, which affects the delay of delivering data to the central server. Adaptive video and image encoding [191, 192] techniques that address the trade-off between available throughput and video distortion can be found effective in meeting the resource contention on the wireless channels to deliver the video with minimum delay end-to-end. In [193], the authors designed a fair energy efficient adaptive video encoding technique for H.264 videos based on tailoring H.264 configurations online according to wireless channel dynamics. In [194, 195] optimal energy efficient techniques for scalable video transport for fixed camera surveillance sensor networks have been developed. The minimization of the total energy spent by the fixed cameras to deliver the video to the central server have been investigated, while addressing the constraints, including video distortion and data loss rate on the network level using a centralized non-linear optimization algorithm. However, these proposed techniques may not be directly applicable in practical UAV systems architecture due to two main reasons. First, the mobility of the UAVs and dynamic network topology makes it challenging to apply techniques based on centralized optimization, assuming the topology and the state of the channels do not change frequently. Second, for the flying UAVs, it may be much more efficient to minimize the end-to-end delay for video transport, rather than minimizing the communication energy, because as we concluded in Section IV-C, the communication energy is negligible compared to the mechanical energy consumed during navigation. Therefore, if UAVs can deliver data faster using more encoding and communication energy, they should prefer that over minimizing energy consumption.

Cross-layer design on the other hand, is also a powerful concept for QoS provisioning, optimizing data transport in different wireless network settings. In cross-layering, different network parameters from different layers are coordinated by the cross-layer controller to achieve the optimal end-to-end performance. In [196] and [197], two optimization frameworks are formulated as video distortion minimization problems under certain resource constraints, typically the delivery delay constraint. The algorithm in [198] considers a joint optimal design of physical medium access control (MAC), and routing layers to maximize the lifetime of energy-constrained WSNs. The work introduced in [199] explores a cross-layer design framework for real-time video streaming that identifies the key parameters to be exchanged between adjacent layers. It proposes adaptive link layer techniques that adjust packet size, symbol rate, and constellation size according to channel conditions to improve link throughput, which in turns improves the achievable capacity region of the network.

In another aspect, With the current growth of several technologies that use wireless devices, a large number of these devices have operated on available bands, thereby introducing congestion on band utilization, leading to spectrum scarcity.

Additionally, the increasing applications of multi-UAVs can lead to spectrum insufficiency. However, cognitive radio has become a potential solution for spectrum scarcity challenges [200].

Cognitive radio can sense and regulate unoccupied bands in the spectrum and dynamically adjust its functioning parameters to use these available bands resourcefully. Consequently, it can hold a practical opportunity for the deployments of several multi-UAVs. On the one hand, the current deployments of UAVs use static spectrum assignment. However, as a consequence of the high mobility of UAVs , licensed spectrum costs, and legal regulations, UAVs do not leverage the licensed spectrum [200]. Therefore, most UAVs use unlicensed spectrum bands. However, with the increase of smart devices and the growth of UAV applications, unlicensed spectrum bands are rapidly becoming congested. On the other hand, cognitive radio can deliver dynamic spectrum access to UAVs to utilize licensed and unlicensed spectrum bands and optimize the performance of UAV networks, such as enhanced packet delivery ratio, better throughput, less delay, and improved reliability [200-202]. Moreover, using cognitive radio technology for UAV systems can minimize energy consumption by reducing the high packet delay that can appear because of packet losses and retransmissions in the congested spectrum bands. Cognitive radio technology can decrease packet losses and retransmissions, as well reduce energy consumption by adjusting and changing network settings and using diverse spectrum bands resourcefully [200, 203]. Another significant application of cognitive radio technology is that it can resourcefully use spectrum bands to provide the needed QoS regarding various application requirements [200]. However, the integration of cognitive radio technology in UAVs poses many design challenges on antenna and multi-UAV collaborations; the issue on the coexistence of different networks and interference controlling among UAVs must be considered [200].

*C. Safety and Security*

The proliferation and low-cost features of such drones pose attractions to many to deploy and use such drones in unexpected manner that can pose privacy, safety or security threats to other humans or network infrastructures. Due to the wireless nature of UAVs and their implementation in open and wide areas, they are subject to be hacked and losing control easily, causing potential harm to people and the environment. For detailed survey on safety and security aspects of multi-UAV systems, the reader is referred to [204, 205]. Also, due to their agility and computational power, drones can be used for hacking, or posing security threats by flying them individually or in-groups inside premises to take control of the network infrastructure [206]. The safety and security issues have triggered many activities to regulate the use of drones [207], while many companies started to focus on building defense anti-drone systems to take control of illegitimate or compromised drones in case they go out of control [208, 209]. Therefore, security considerations should be considered when forming multi-drone networks. First of all, the area of interest must be permitted for physical presence of personnel, which lowers the chance of malicious actions [210]. On one hand keeping the transmitting power of the drone system could help the network from being eavesdropped but on the other hand, it could affect the connectivity of the UAVs to the system and thus be a cause of failure. To overcome this issue, algorithmic solutions for UAV networks should provide autonomous and self-configuring mechanisms.

Traditional security goals, such as authentication, secrecy, and non-repudiation, have been well studied in wired networks and users are usually aware of the risks due to the lack of security in such systems. In civil UAV networks, the adversary models and the type of attacks follow the same concepts [211], yet, security was not the first priority for the designers of such systems since many proposed protocols are not secured and often leak information sent over the network.

Although adding redundant UAVs to the system could lower the efficiency of the whole network in terms of consumed energy and hardware but, it could guarantee the network reliability. Enhancing the network operation is reachable by smartly adding the drones when it is deemed necessary. For example, adding a UAV at a right place could help covering an area in shadow in terms of wireless connectivity or, could help monitoring a target from additional direction which has not been covered previously. Also, redundant UAVs could be added to replace some existing UAVs so as to increase the flight time of the whole swarm.

To enhance the safety of the drone, failure detection should be investigated. Failure detectors have been investigated to detect crashes among networks for a long time. For example, in the indoor UAV testbed in [46], the authors developed a simple failure detection mechanism that allows UAVs to detect the boundary of the test area through marked (colored) lines, and land safely once they fly past these lines. Authors in [212] introduce unreliable failure detector to circumvent the impossibility of distributed consensus with one faulty process in an asynchronous distributed system proved by [213] . Many other failure detectors have been proposed in [214, 215]. However, the application of such failure detection algorithms for dynamic environments with changing UAV topology needs to be investigated.

*D. Lessons learned*

Network connectivity challenges of UAV are critical factors to be considered while designing multi-UAVs for CPS applications. Unlike several other wireless networks, UAV network topologies are dynamic with several nodes and connections because the positions of the nodes are dynamically changing. On the basis of the different CPS applications, the UAVs need to accelerate with different velocities; consequently, this acceleration can affect connectivity with other UAVs, thereby causing the irregular connection and unstable link. Some properties of the design challenges of network connectivity are based on several factors, such as the type of the mission and environment; in addition, the needed solutions may differ accordingly. The dynamic nature of the UAV network, the vanishing nodes, and unstable link can lead to vital issues for the designer in planning methods that tackle these issues beyond the regular ad-hoc mesh networks. Moreover, the implementation of routing protocol cannot be straightforward in such networks [8, 14].

Another main design challenge is to provide the users' sessions seamlessly and to consider delay tolerance, energy limitations, communication range, and controlling and coordination requirements [8, 14]. Other important challenges that should be considered while designing the Multi-UAVs in CPS applications are as follows. Using multi-UAVs in CPS applications can extend the attack surface of the systems that can include physical and cyber vulnerabilities. Moreover the introduction of internet of things (IoT) will enhance the applications of UAV systems [216]; on the other hand, the integration into IoT systems can enlarge the vulnerability surfaces of UAVs systems, because the IoT systems have several attack surfaces [217].Therefore, with the wide range of UAV applications, security is expected to be a high priority while designing UAVs. However, [218] showed that UAVs are not as secure as expected; this study reported that man-in-the-middle attack can be launched and the UAV can be controlled even with long distances from the main controller. Several researchers have identified several security vulnerabilities against UAVs. Authors of [219] discussed cyber-attacks that potentially affect the confidentiality, integrity, and availability of UAV systems. Authors of [220] considered the vulnerabilities that can be exploited in the communication links between ground control and UAVs. Authors of [221] proposed an encrypted radio control link for securing communication links among UAV systems. Authors of [31] examined the security of UAV delivery systems against cyber-physical attacks.

However, exploring the literature provides that even with the security of UAVs only few studies emphasized examining and tackling the underlying design security challenge. Further research, investigations, and solutions in this field are required in the future. Table 10 maps UAV network connectivity, QoS, and security design challenges and requirements with CPS applications.

TABLE 10

MAPPING OF UAV NETWORK CONNECTIVITY, QOS, AND SECURITY USE CASES, DESIGN CHALLENGES, AND REQUIREMENTS WITH CPS APPLICATIONS

| Applications | Use Cases, Design Challenges and Requirements |
|---|---|
| **Transportation** | • Designing UAVs system with secure and reliable connectivity has significant role for several use cases such as tracking moving vehicles at diverse speed, integrating UAVs into IoT system to accomplish smart transportation system<br>• Intelligent transportation includes moving target; consequently, moving UAVs for effective tracking can unstable connection links.<br>• Deployment of UAVs in intelligent transportation leads to a high mobility of the UAVs and dynamic network topology that can affect the delay of delivering data to the central server.<br>• Intelligent transportation used IoT systems to connect effectively all the components in the systems and to control remotely and interact with systems; however, utilizing IoT systems can increase the vulnerability surfaces of UAV systems by potentially involving attacks from the internet.<br>• Current flying ad-hoc networks are vulnerable. |
| **Constructions and Infrastructure inspections** | • Secure and reliable connectivity are crucial for use cases such as establishing reliable long-distance infrastructure inspections, and effective and real-time construction progress tracking.<br>• Connection links may be unstable particularly in using UAVs for infrastructure inspections in hard-to-reach areas.<br>• Security issues may be experienced particularly in unattended areas<br>• Using UAVs for infrastructure inspections has high mobility that can affect the delay of delivering data to the central server. |
| **Surveillance** | • Designing UAVs system with secure and reliable connectivity is vital for many use cases such as border surveillance, and disaster surveillance.<br>• Surveillance applications involve sensitive data, and any data leaking can be critical<br>• Surveillance applications of UAVs moving at different speed may cause unstable connection links.<br>• Using UAVs for surveillance applications are vulnerable against privacy breaching |
| | • UAVs for surveillance applications most likely are integrated to use IoT systems. IoT has large vulnerability surfaces that can increase the vulnerabilities of using UAVs for surveillance applications<br>• Security issues are experienced in unattended environments. |
| **Delivery of goods** | • Secure and reliable connectivity is important for uses case such as providing practical delivery systems, and developing UAVs for goods delivery using IoT systems that allow to track and monitor the delivery process anytime, anywhere.<br>• Deployment of UAVs in the delivery of goods can affect the delay of sending data to the central server.<br>• Using UAVs for the delivery of goods is effective integrated with IoT systems to track and monitor the delivery system anywhere anytime effectively; however, IoT systems can increase the vulnerability surfaces of UAV systems. |
| **Wireless & cellular systems** | • Designing UAVs system with secure and reliable connectivity is the key success to use UAVs as flying base station that can deliver a wide range of wireless networking applications.<br>• Effective security schemes must be employed to protect the UAVs from the so-called ghost control scenario; this scenario is the unauthorized control of UAVs through spoofed control or navigation signals [33].<br>• An efficient deployment of multi-UAVs for wireless network applications requires complex various tradeoffs involved in various UAV deployment scenarios to be considered for providing optimal wireless coverage [222, 223]. |
| **Medical and Healthcare Systems** | • Secure and reliable connectivity is extremely compulsory for all healthcare use cases, given the human life as the high priority cannot be compromised.<br>• An advantage of using UAVs for medical and healthcare systems is to reach locations that are difficult to reach before. However, UAVs in such environment may have unstable connection links that can compromise their missions.<br>• Medical systems should be a secure and real time response system. However, the trade-off between security and real time response in medical systems is a challenge. Providing highly secure systems may lead to complex systems. In other hand, providing highly flexible systems for real time response may lead to less secure systems. |

## VII. FLIGHT CONTROL

A trustworthy model that can precisely capture the flight dynamics is vital to design effective autonomous flight control schemes. The aim of designing fully autonomous flight controls is to be capable to operate in different environments and accomplish several required missions which is essential to close the circle in CPS systems through controlling the UAVs as a result of environment dynamics; therefore, UAVs must prove the practicality and operability in real flight examinations.

Researchers of [224] mentioned that the flight control system mainly involves three parts, that is, the kernel control, command generator, and flight scheduling. First, the purpose of kernel control law is to ensure the asymptotic stability of the UAV motion through air. Second, the purpose of the command generator is to generate a set of rules or references for the kernel control. Third, flight scheduling function is to create a set of flight arrangements for the mission.

Generally, two fundamental schemes control the flight of UAVs, namely, through controlling altitude and through controlling heading and velocity [225]. Controlling altitude is to manage the UAV to maintain flying at required altitudes. Controlling heading and velocity is to manage the UAV to fly through specified and required paths. Consequently, to accomplish the fundamental control schemes, several strategies, proportional–integral–derivative controller (PID) controller, learning-based schemes, and linear flight controllers, have been proposed.

### A. Proportional–Integral–Derivative (PID) Controller

PID method is one of the most commonly used method among commercial controllers for UAVs. In practice, PID automatically uses precise and responsive adjustment to control functions. In PID, the controlling parameter is adjusted to tune the current state of the UAVs with reference flight path and kept continuously tuned throughout the flights. PID-based controllers are popular, because they are easy to be used in small UAV systems. However, PID controllers have limitations in optimality and robustness. Besides, their parameters are also difficult to tune under some circumstances [225].

### B. Learning-based methods

Learning-based methods for controlling UAVs have become popular recently. The traditional control model can only be used within a specific environment and is mostly capable to adapt to modifications or change to unexpected circumstances or aggressive environments. Therefore, to cope with these challenges, learning techniques are adopted. These techniques can overcome such issues and learn from real-world experience to provide adaptive control schemes. The important advantage of learning-based control methods is their ability to learn from real world scenarios and behave accordingly. Recent advances in learning methods, such as deep learning, allow to learn a pattern effectively from raw data under complex environments.

For the flight control of UAVs, the inputs of the learning can be images, light imaging detection and ranging sensor data, or both [226]. Learning the pattern from such inputs can allow to learn proper controlling schemes even under unknown circumstances. For instance, recent studies have shown the effectiveness of deep learning algorithms to map images to high-level control instructions (e.g., turn left, turn right, rotate left, and rotate right) [226-228]. In study [66], a deep learning model (ConvNet) was utilized to learn a controller approach that mimics an expert pilot's selection of actions. The authors validated the performance of proposed systems by showing the performance of the systems using a real-time investigation in various indoor locations; they also proved the significance of deep learning approach application in micro UAVs for a real-time autonomous indoor navigation.

### C. Linear Flight Controllers

Given that most UAVs operate in complex models, modeling techniques are typically nonlinear models. However, because precisely modeling such environments is difficult, traditional methods for controlling flight and early efforts to accomplish autonomous UAVs flight are based on linear controllers, such as LQR, linear quadratic Gaussian (LQG) /loop transfer recovery (LTR), LTR, and $H\infty$. A linear model can be applied to model UAV dynamics. A hybrid scheme of an LQG controller and Kalman filter is used to accomplish an improved altitude control performance [229]. In another study, LQG\LTR multivariable control is used to control the UAV attitude in the existence of noise and disturbance [230]. Table.11 presents comparison among common flight control methods.

TABLE 11

COMPARISON AMONG COMMON FLIGHT CONTROL METHODS

| Flight control methods | Advantages | Disadvantages |
|---|---|---|
| PID | • Easy to understand, design, and implement<br>• Can be integrated to high-level control approaches<br>• Requires less processing and memory resources | • The parameter tuning is complex.<br>• If the payload is changed, then the parameters should be retuned.<br>• Stability is sensitive to external disturbances. |
| Learning-based methods | • Utilize robust methods that can provide accurate control, such as deep learning<br>• Can integrate learning methods into AR technology<br>• Provide interactive interface for good controlling approaches<br>• Can provide a good stability compared with others in complex unknown environments. | • The methods have high computational costs and may suffer from delay.<br>• The learning methods need high processing and memory resources.<br>• The learning methods required huge data for training before they can provide a high accuracy performance. |
| Linear flight controllers | • Easy and simple to design and implement<br>• Requires less processing and memory resources | • The controllers suffer from accuracy issues.<br>• With the increase of the applications of UAVs in CPSs, the operational environments become more complex and involve unknown parameter. |

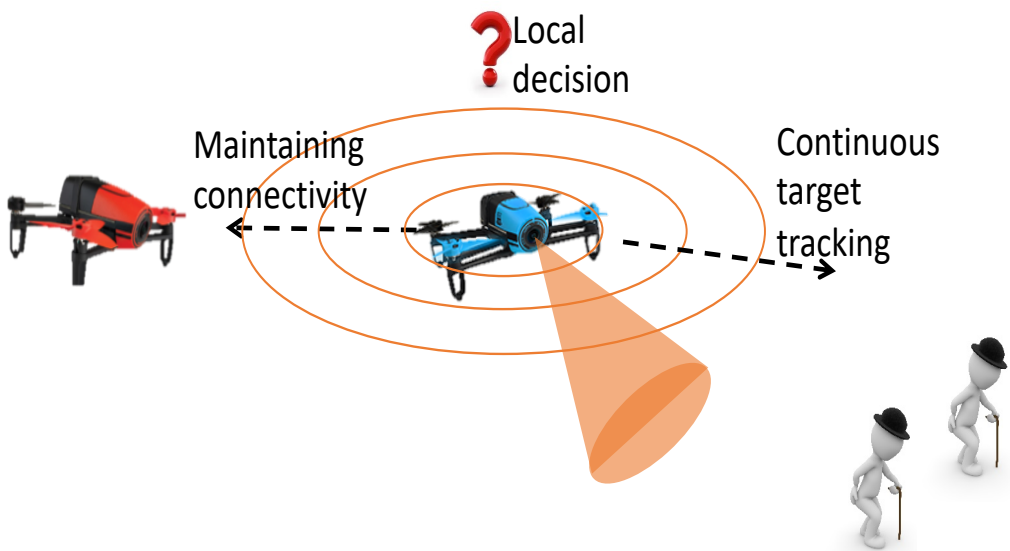

Fig. 23. Distributed command and control with conflicting objectives

## VIII. FUTURE IDEAS AND MOTIVATIONS

Despite all the efforts to propose reasonable solutions to address the design challenges discussed above for multi-UAV systems, significant amount of unsolved difficulties is waiting for an effective remedy. Besides, the development of such systems could bring about a more safe and reliable life to the human beings along with a promising opportunity for the individual researchers and industries to further work on this area and make it as practical and easy to implement as possible. In the following, we discuss some of the potential research directions for multi-UAV systems along the same lines of the challenges discussed before. It can be seen that a common theme is shared amongst all these research directions; that is addressing the design challenges discussed above jointly will either be Inevitable or will be required to enhance the system performance. Future directions of UAV design challenges for various CPS applications are summarized at the end of this section in Table.12.

### A. Distributed UAV architectures addressing multiple objectives

The architectures we discussed in section II cover mainly multi-UAV systems with continuous navigation through a centralized command and control server (CCS), which provides at least global high-level control to UAVs to monitor targets/areas in real-time. Hence, continuous connectivity between all drones and the CCS is a condition to guarantee continuous navigation. Autonomous swarms also tend to nominate a designated leading drone to perform the function of the CCS, although this scenario is more challenging because of the limited computational capacity of the mobile CCS. In both cases, however, centralized navigation and control affects the scalability of the system, limiting the number of drones participating in the swarm mission, so decentralizing navigation and control through peer-to-peer, clustered topologies, or distributed autonomous drone navigation is an interesting area that is required to address many new challenges. Some of these new challenges include, but not limited to, maintaining connectivity within the swarm while conducting a unified mission, and addressing the tradeoff between coverage/tracking, while maintaining connectivity. Figure 23 illustrates this issue, showing a UAV trying to make a local decision with two conflicting objectives to maintain wireless connectivity with other drones, while maintaining continuous target tracking, which is challenging with the lack of global view. In general, designing distributed architectures with distributed command and control may involve addressing a subset or all of the following conflicting objectives, which is a promising area of research:

- Maintain network connectivity
- Maintain all targets within coverage
- Distributed decision making with global swarm state
- Secure communication within the UAV swarm
- Maximize the viewing angle while avoiding occlusion between targets
- Maximize the life time of a swarm
- Minimize the communication energy for delivering sensor information to the central server
- Minimize the service or event detection time
- Minimize the distance between the camera and all targets in the view Minimize the end-to-end delay for real-time monitoring
- Maximize quality of images and videos

Naturally, distributed algorithms and techniques such as greedy, genetic or game-theoretic approaches would be handy in such scenarios because of their flexibility in distributed architectures. For example, clustered swarm formation can be mapped into leader/followers Stackelberg game [231], where the cluster head is a leader who makes decision based on maintaining localized view of the drones in its vicinity, then the other drones in the cluster would follow consequently. On the other hand, some early works tried to address some of the challenges above such as global swarm state, and secure communication through leveraging Blockchain technologies [232, 233], However, such architectures may not work efficiently for real-time monitoring and tracking, and hence other innovative solutions are still needed.

### B. Coverage and tracking through vision-based detection

In section III, we discussed target coverage and tracking where the locations of the targets are either known deterministically [79], statistically through prediction models [100], or at least coordinated between the UAVs in a distributed manner [92]. In some applications such as crowd management and crowd control, knowing the locations of the large density targets is a daunting process, and hence, we anticipate the emergence of a new class of solutions that focus on evolutionary crowd monitoring through strategic area coverage. The main idea is to divide the area into virtual shapes depending on the density of the crowd in this location, which can be detected by the UAV cameras through vision-based techniques [234], and then assign and coordinate dynamically the number of UAVs required to cover each location depending on the target crowd density. Several challenges need to be addressed in this case to predict incremental movements of targets within different areas without moving the UAVs, and targets handover amongst UAVs to achieve optimal coverage and tracking in real-time. Some of these challenges can be addressed through auction-based techniques, where UAVs can bid on covering incrementally certain areas based on optimizing a utility function combining multiple objectives, e.g., cover high density areas, energy consumption, etc.

On the other hand, adding targets' dimensions and directions is indeed another promising area of research, which has been validated by the fact that UAVs can fly in low altitudes and hence can provide rich information about such targets. The early works discussed in section III-B1b has motivated the need for future work related to modeling simplified targets to represent directional $3D$ targets (see Figures 13 and 14), and directional monitoring of crowd to maximize visual information and quality, etc.

### C. Path and trajectory planning

New techniques should be developed for coverage and tracking of targets, while optimizing the paths the UAVs take to perform these functions with minimum number of UAVs and minimum energy budget in order to maximize the UAV flight times. The techniques discussed in section IV mainly consider end-to-end paths with straight lines in one or two $2D$ planes to simplify the end-to-end path planning, while avoiding collisions amongst UAVs. Since the general coverage problem with most of its variants are proven to be NP-Hard [39] , [57], all work in this area is based on developing heuristic algorithms that require low complexity, especially if target tracking is part of the objective. Planning complex (i.e., irregular) paths that UAVs go through to cover key areas, and objects onsite may be deemed useful to minimize the aerial mechanical energy consumed for navigation, while insuring that the UAVs can fly in an obstacle-free plane, and avoid collision amongst each other. To do so, one could think of forming a multi-objective optimization problem and create a model that describes a specific application then try to solve it for some goals and under some critical assumptions.

### D. Networking and cross-layer design

The deployment of traditional sensor networks in harsh energy sites such as oil and gas environments may not be feasible due to interrupted wireless communications, and constant human intervention for maintaining such sensors to guarantee continuous operations. Large pipelines and oil tanks require, however, regular monitoring through measuring wear and tear to efficiently provide preventive maintenance for such important infrastructure, which may not be feasible manually. It is roughly estimated that the unexpected shutdown of one oil production line may cost over 10 million dollars per one single day. Therefore, multi-UAV systems will help provide effective and innovative ways of monitoring such important infrastructures without permanently deploying sensors intrusively to that environment. However, new connectivity models may be needed for topology control in such harsh environments to guarantee efficient communication amongst mobile drones. Also, augmenting the techniques in [187, 188] [189, 190], through combining topology control problems with target coverage and tracking is indeed key for improved performance of the multi-UAV system.

Also, leveraging MEC for video analysis and processing in a cross-layer design to optimize end-to-end delay and energy consumption of multi-UAV systems is key. The work in [194, 195] provides good example of combining application level characteristics, i.e., video quality, with resource allocation in UAV networks. However, these techniques cannot be applied directly to mobile cameras on UAVs due to high complexity, and hence, extending these models for drones' MANET is a natural research direction.

### E. Safety and security

Due to the wireless nature of UAVs, their limited computational capabilities and their deployment in open areas (urban area, open sky, etc.), they are prone to malicious attacks. Thus, measures should be taken to ban such kinds of attacks. Besides this, there have to be careful considerations that make UAVs connected to the system during their life time to prevent them from unwanted and abnormal actions. UAVs could use failure detectors to detect crash among networks. Developing powerful intrusion detection algorithms could also diminish collision between the drones in case one or more UAVs are compromised. Therefore, generally, the area of safety and security for multi-UAV systems is rather unexplored, and hence, discussions in section VI-C introduce many problems without efficient solutions such as drone regulations, privacy techniques, drone defense systems, robust intrusion detection techniques in UAV swarms, forensics for post-event analysis, etc. Drones are perceived to be privacy intrusive due to their ability for video capture in unexplored angles and areas, and with UAV swarms, we can add multi-view ability, which is a very crucial issue [205]. Therefore, new on-board techniques may be required for each drone to insure privacy while capturing videos. Other approaches to insure privacy through defense systems that form virtual bubbles around key areas to divert drones from intruding the privacy of human or infrastructure [208, 209]. Such systems require the detection, identification, tracking, and possibly jamming UAVs that pose safety or security threats. On the other hand, forensics solutions are also needed to insure important events are logged by the multi-UAV system to allow for post-event accountability.

### F. Machine learning

Machine learning in general and its sub-field deep learning in particular recently reached a remarkable level of supporting practical applications that can effectively utilize these algorithms to enhance the autonomous navigation ability of UAVs as well as optimize UAVs performance in many vital applications, such as object detection and tracking, path planning and navigation, reactive obstacle avoidance, and aggressive maneuvers. [226, 235-238].

On the one hand, commercial evolution of UAVs has already started manufacturing the Nano-scale, which introduces a few centimeters in the diameter of UAVs with the less weights [235, 239]. However, current Nano-UAVs have limited capabilities to implement computational methods for onboard vision-based autonomous navigation. On the other hand, developing lightweight algorithms and improving onboard computational capability of these Nano-UAVs are desirable future directions that can make UAVs suitable for several CPS applications. Besides, computational algorithms that enable the machine intelligence in UAVs can be implemented in remote resource power-unconstrained machines. However, such deployments have challenges that should be tackled in the future, such as communication latency, bandwidth limitation, and noise [240].

Besides the several potential uses of machine learning algorithms for various applications of UAVs for CPS application, machine learning can be used for potential collaboration between multi-UAVs to achieve a mission or avoid the collision intelligently in multi-UAV congested with flying UAV deployments. Another future direction can be to enhance the use of machine learning for spectrum sensing and white space identifications in cognitive radio, massive MIMO channel estimation, and intelligent energy management [1].

TABLE 12

FUTURE DIRECTIONS OF UAV DESIGN CHALLENGES FOR VARIOUS CPS APPLICATIONS

| Applications | Future Directions |
|---|---|
| Transportation | • Utilizing the data from various sources of multi-UAV systems provides rich multi-view images and videos.<br>• Swarm intelligence methods that can effectively provide cooperation between UAVs and support and utilize fusion of the data from various UAVs and sensors will hold practical importance for transportation systems.<br>• Minimizing the event detection time is needed for fast tracking mechanisms.<br>• Maximizing quality of images and videos is required to get useful insightful information required in transportation system.<br>• Minimizing the end-to-end delay is required for real-time monitoring.<br>• Methods that ensure protecting privacy, such as vehicle location, movement, and vehicle identification, should be provided [1].<br>• UAVs that operate in secure environments should be insured, specifically, smart transportation systems that include IoT devices that may open the door to several vulnerabilities from the Internet and the cyber world. |
| Constructions and Infrastructure inspections | • If the construction and infrastructure need long distance, then the inspection coverage needs to maintain network connectivity, thereby ensuring the required area of inspection is within coverage.<br>• Using the image from various multi-UAV systems provides rich multi-view images and videos.<br>• Developing effective 3D representation schemes is needed to enhance the analysis and improve the monitoring and inspection process.<br>• If multi-UAV systems are integrated with technologies such as AR, then such integration can lead to potential applications in constructions and infrastructure inspecting, delivering, and assisting processes. |
| Surveillance | • The surveillance application may include unpredictable target that may move along unpredictable paths, tracking systems, and detection methods with high speed and accuracy are required.<br>• Effective data fusion schemes can provide diverse data about the area to be scanned and enable comprehensive scanning.<br>• The viewing angle should be enhanced while avoiding occlusion among targets, particularly in wide and urban areas.<br>• Machine and deep learning algorithms can have potential applications for path planning and target tracking under uncertainty mode.<br>• A future work for this application should ensure the privacy preservation of the scanned area, particularly in urban areas or areas occupied by humans.<br>• Effective security methods should be developed, particularly if the UAV is part of the IoT systems. |
| Delivery of goods | • Safe landing and delivering schemes should be investigated and developed. Machine learning may provide methods that support safe landing and delivery of goods.<br>• The design of UAVs for the deliveries of goods should be cost effective compared with other traditional delivery systems.<br>• Designing and developing effective methods should ensure that goods reach the exact delivery address.<br>• Security challenges should be considered when designing UAVs for the delivery of goods systems.<br>• The effect of various weights and sizes of payload on UAVs flying stabilities as well as other flying factors should be considered when designing UAVs for delivery of goods systems |
| Wireless & cellular systems | • Realistic channel models that reflect actual world measurements are needed [17, 241].<br>• An effective network planning that can provide an optimal balance among the minimum number of UAVs is required to deliver full coverage for a specified terrestrial area effectively [14].<br>• When designing, considering the optimal management of different factors, such as bandwidth, energy, transmit power, UAV's flight time, and number of UAVs, is required [14].<br>• Interference mitigation methods should be designed successfully for massive UAV deployment scenarios [17]. |

| Medical and Healthcare Systems | • Efficient flight controls and path planning schemes are required to provide stable flight missions. |
| | • The security of UAV and its connection during the flight should be considered in using UAV for healthcare applications. |
| | • Responsive analytical methods are needed for emergency cases in health search-and-rescue applications. |
| | • Methods that ensure protecting the privacy, such as patient location, identification, and information, should be provided. |

## IX. CONCLUSIONS

In this paper, the design challenges of Multi-UAV in Cyber-Physical applications have been classified then elaborations on each were given. Difficulties and remedies for each of the challenges were studied to show how each of them could affect the performance of the multi-UAV system. Besides, various algorithms of fixed and mobile target coverage and tracking were investigated along with a comprehensive comparison between them regarding several metrics like complexity, the fraction of uncovered area and the number of surveillance cameras. At the end, future ideas and motivations provided the reader with a promising perspective of the next generation of multi-UAV systems, addressing many issues related to scalable and decentralized architectures, wireless connectivity and topology control in large UAV networks, coverage and tracking of unpredictable and dimensional targets, path planning, safety and security, and many more.

**Acknowledgements:** This publication was supported by Qatar university Internal Grant No. QUCP-CENG-2018\2019-1. The findings achieved herein are solely the responsibility of the authors.

TABLE 13: ACRONYMS

| AoV | Angle of view |
| --- | --- |
| ADS-B | Automatic Dependent Surveillance-Broadcast |
| CF | Cluster-First (CF) |
| CCS | Command and Control Server |
| CPSs | Cyber-Physical Systems |
| DbeG | Drone-be-Gone |
| DCNN | Deep Convolutional Neural Network |
| D-Smp | Dual Sampling |
| EH | Energy Harvesting |
| EGPSO | Evolutionary Game Particle Swarm Optimization |
| EPU | External Processing Unit |
| FoV | Field of View |
| FCM | Fuzzy C-Means |
| FC | Fuzzy clustering |
| GA-ACO | genetic Algorithm and ant Colony Optimization |
| GPS | Global Position System |
| IoT | Internet of Things |
| LIFA | Local Incremental Fuzzy Algorithm |
| MG | Manhattan Grid |
| MDPs | Markov Decision Processes |
| MAC | Medium Access Control |
| MANETs | Mobile ad-hoc Networks |
| MEC | Mobile Edge Computing |
| NIR | Near Infrared |

| NBO | Nominal belief-state optimization |
| --- | --- |
| OEM | Original Equipment Manufacturer |
| PID | Proportional–Integral–Derivative |
| POMDPs | Partially observable Markov decision processes |
| PSO | Particle Swarm Optimization |
| PV | Photovoltaic |
| PFA | Predictive Fuzzy Algorithm |
| PIFA | Predictive Incremental Fuzzy Algorithms |
| PHs | Pythagorean Hodographs |
| QoS | Quality of service |
| RWP | Random Way Point |
| RPG | Reference Point Group |
| RSS | Receiver Signal Strength |
| SRS | Satellite Remote Sensing |
| SNs | Sensor Networks |
| SCSC | Simple Cover-Set Coverage |
| SWIPT | Simultaneous wireless information and power transfer |
| SSKCAM | Smart Start K-Camera Clustering |
| SGD | Stochastic Gradient Descent |
| TOP | Team Orienteering Problem |
| UAVs | Unmanned AiVehicles |
| UnA | Up & Away |
| VANETs | Vehicular ad hoc networks |
| WPT | Wireless power transfer |
| WSNs | Wireless sensor networks |
| WVS | Wireless video surveillance. |